%% file: main.tex
\documentclass{article}

\PassOptionsToPackage{numbers, compress}{natbib}

\usepackage[main, final]{neurips_2026}

\usepackage[utf8]{inputenc}
\usepackage[T1]{fontenc}
\usepackage{hyperref}
\usepackage{url}
\usepackage{booktabs}
\usepackage{amsfonts}
\usepackage{nicefrac}
\usepackage{microtype}
\usepackage{xcolor}
\usepackage{graphicx}
\usepackage{subcaption}
\usepackage{sidecap}
\usepackage{amsmath}
\usepackage{multirow}
\usepackage{enumitem}
\usepackage{threeparttable}
\usepackage{todonotes}
\usepackage{array}

\newcommand{\abar}{\boldsymbol{\bar{\alpha}}}
\newcommand{\ba}{\boldsymbol{\alpha}}
\newcommand{\bb}{\boldsymbol{\beta}}

\title{Everything at Every Scale: Scale-Invariant Diffusion with Continuous Super-Resolution}

\author{
Zixin Jessie Chen$^{13}$\thanks{Equal contribution}\;\;\thanks{Corresponding author}\quad
Zhuo Chen$^{134}$\footnotemark[1]\quad
Archer Wang$^{23}$\quad
Jeff Gore$^{1}$\quad
William T. Freeman$^{23}$\\
\textbf{Congyue Deng}$^{2}$\quad
\textbf{Marin Solja\v{c}i\'{c}}$^{13}$
\vspace{0.05in}
\\
$^{1}$Department of Physics, Massachusetts Institute of Technology \\
$^{2}$Department of EECS, Massachusetts Institute of Technology \\
$^{3}$NSF AI Institute for Artificial Intelligence and Fundamental Interactions \\
$^{4}$Institute for Data, Systems and Society, Massachusetts Institute of Technology\\
\texttt{\{jzxchen,chenzhuo,archerdw,congyued,billf,soljacic\}@mit.edu}
}

\begin{document}
\maketitle

\vspace{-20pt}

\begin{figure}[h]
    \centering
    \includegraphics[width=1.0\linewidth]{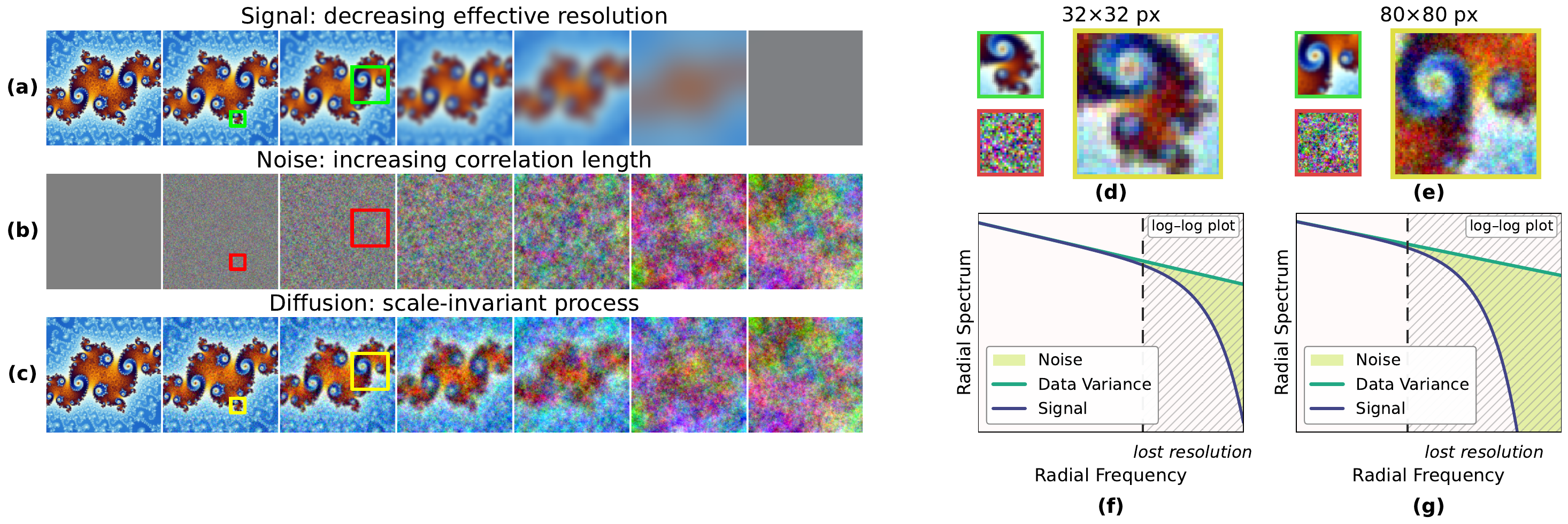}
    \caption{
    Conceptual illustration of our SKILD on a self-similar fractal image. During the forward process: (a)
    effective signal resolution decreases; (b) the pixel-space correlation length of the injected noise increases; and (c) for a
    self-similar field, the process respects the same frequency-space power spectrum across stages.
    (d-e) A smaller early-time patch is statistically similar to a larger
    late-time patch. (f-g) Radial power spectrum at
    corresponding early and later stages. Gray slashed regions indicate modes below the signal-to-noise
    ratio (SNR) threshold, where resolution is effectively lost.}
    \label{fig:concept}
\end{figure}

\begin{abstract}
    Creating images from noise is image generation; reconstructing fine
    details from coarse inputs is super-resolution. Despite their practical
    differences, both can be understood as reversing information loss across
    scales. We introduce \textbf{SKILD}, a \textbf{S}cale-invariant
    \textbf{K}-Space \textbf{I}mage \textbf{L}earning \textbf{D}iffusion
    model that unifies generation and continuous super-resolution within a
    single unconditional framework. Both natural images and critical physical systems exhibit scale invariance, and we leverage it to design a forward process that attenuates
    image content from fine to coarse scales while injecting
    spectrum-matched Gaussian noise, making scale an explicit coordinate of
    the diffusion dynamics. The same trained reverse process performs
    generation and continuous super-resolution
    by varying only the starting timestep: \textit{no task-specific architecture, no conditioning branch, no classifier-free guidance, no retraining per
    scale factor}. Empirically, SKILD reaches FID $2.61$ and Inception Score
    $9.65$ on unconditional CIFAR-10, performs $2\times$--$8\times$
    super-resolution on ImageNet from a single unconditional checkpoint while outperforming conditional models across perceptual metrics, and reconstructs
    critical Ising models whose connected four-point correlations closely track the ground truth.
\end{abstract}

\input{sections/1_intro}

\input{sections/2_related_works}
\input{sections/3_prelim}
\input{sections/4_theory}

\input{sections/5_experiment}

\input{sections/6_scientific_data}

\input{sections/7_conclusion}

\bibliographystyle{naturemag}
\bibliography{refs}

\input{sections/10_appendix}

\newpage
\input{checklist.tex}

\end{document}

%% file: sections/1_intro.tex
\section{Introduction}\label{sec:intro}

Scales in images have long been a subject of study in computer vision. Across
different scales, images share recurring structure. A zoomed natural image
still looks natural, and some natural objects are themselves self-similar, with
textures, edges, and structures recurring at different scales. Statistically,
this regularity is reflected in natural-image power spectra, which follow
approximate power laws over wide frequency ranges
\cite{field1987relations,ruderman1994statistics,vanderschaaf1996modelling,simoncelli2001natural},
a signature of approximate scale invariance. The same concept has been studied
in parallel in physics, where critical systems display similar scale-invariant
behavior, made formal by the renormalization group
\cite{wilson1974renormalization,belavin1984infinite}. This physics perspective
also points to a natural way of organizing the transformation between an image
and pure noise. Can we take advantage of this scale invariance in diffusion?
Rather than corrupting all scales at once, one can erase them in order, one scale
at a time. Diffusion, framed in this way, becomes a denoising process respecting self-similarity across scales.

Such a denoising process across scales is, by construction, a progressive
super-resolution. At each backward step, finer scales are added back, and
running the full reverse process from pure noise produces an image scale by
scale. This unifies generation and super-resolution into a single framework.
Generation from noise is the extreme case of super-resolution in which the
input contains no signal at all; super-resolution is the same reverse process
initialized from an intermediate state in which coarser scales have survived.
Both are reverse coarse-graining problems, distinguished only by where the
reverse process begins.

We realize this idea with \textbf{SKILD} (\textbf{S}cale-invariant
\textbf{K}-Space \textbf{I}mage \textbf{L}earning \textbf{D}iffusion), a
diffusion model whose forward process corrupts images one scale at a time, from
finest to coarsest. Two design choices make this concrete. First, the forward
process attenuates high-frequency content before low-frequency content. Second,
the noise added at each step carries the spectrum of the dataset itself rather
than being white noise, so the model learns to remove noise that  
statistically resembles the data it learns to generate. Together, these two choices
make every intermediate state a coarse-grained, noisy version of the
original image in a self-similar manner. 

Our contributions are as follows.
\begin{itemize}[leftmargin=1.2em, labelsep=0.4em, itemsep=0.25em, topsep=0.25em]
    \item We propose SKILD, a scale-invariant diffusion framework that
    unifies unconditional generation and continuous super-resolution within a
    single reverse process. A single, unconditional architecture handles both
    tasks, replacing what would otherwise be a stack of task-specific
    architectures, conditioning branches, classifier-free guidance, and
    per-scale retraining.

    \item On unconditional CIFAR-10 \cite{krizhevsky2009learning}, SKILD is competitive with state-of-the-art
    diffusion models and achieves the strongest sample quality among
    frequency-informed diffusion models.

    \item One trained SKILD checkpoint performs continuous super-resolution at
    any factor, which we test on ImageNet \cite{deng2009imagenet} between $2\times$ and $8\times$. At $4\times$
    super-resolution on ImageNet-$256$, the same model outperforms strong diffusion-based conditional
    super-resolution baselines on multiple perceptual quality metrics.

    \item Evaluations on a scientific dataset generated using a critical Ising model show that SKILD reproduces explicit self-similar statistics
     while a strong diffusion-based conditional super-resolution baseline fails.
\end{itemize}

%% file: sections/2_related_works.tex
\section{Related Works}\label{sec:related_works}

\textbf{Scale invariance and self-similarity.}
Scale-space theory analyzes images through continuous smoothing and identifies
Gaussian convolution as the canonical linear scale-space operator
\cite{witkin1984scale,koenderink1984structure,lindeberg1994scale,
lindeberg1998feature,lowe2004distinctive,burt1987laplacian}. Natural-image
statistics show approximate power-law spectra across scales
\cite{field1987relations,ruderman1994statistics,vanderschaaf1996modelling,
simoncelli2001natural,olshausen1996natural}, while renormalization group
theory describes how distributions transform under coarse-graining and rescaling \cite{wilson1974renormalization}. These ideas motivate our
forward process: attenuation from fine to coarse scales in frequency space,
with noise covariance matching the dataset distribution.

\textbf{Diffusion models across scale and frequency.}
Diffusion models learn to reverse a noising process \cite{ganguli2015deep, ho2020denoising, song2020score}, with various samplers and schedules \cite{nichol2021improved, dhariwal2021diffusion, karras2022elucidating, chen2023importance}. Several lines of work connect diffusion to multi-scale
structure: cascaded and relay models compose resolution-specific conditional stages \cite{ho2022cascaded,teng2023relay}; other work connects diffusion to renormalization-group flows, optimal transport, or inverse heat dissipation \cite{cotler2023renormalizing, rissanen2022generative,masuki2025generative, sheshmani2025renormalization}. A separate line uses Fourier or wavelet structures to improve controllability, efficiency, or inductive bias \cite{guth2022wavelet, phung2023wavelet, ning2024dctdiff, falck2025fourier, yu2025frequency, moser2024waving, gao2024frequency, friedrich2024wdm}. Recent works have also explored image generation as progressive super-resolution in pixel space, replacing additive noise with structured degradations or multi-scale reconstruction processes \cite{bansal2023cold, tian2024visual}. Unlike these approaches, SKILD explicitly utilizes self-similarity in frequency space, where the forward process continuously attenuates image statistics from fine to coarse modes. As a result, a single reverse process supports both unconditional generation and continuous super-resolution without conditioning or guidance.

\textbf{Super-resolution.}
Beyond classical priors
and feed-forward neural methods
\cite{freeman1999learning,dong2015image,lim2017enhanced,wang2018esrgan,
zhang2021designing,liang2021swinir,wang2021real}, diffusion-based methods often rely on additional conditioning from low-resolution images \cite{saharia2022image,li2022srdiff,yue2023resshift,wang2024sinsr,
kawar2022denoising,luo2023image,liu20232}. SKILD requires no extra 
conditioning: the low-resolution input is an intermediate state of the
model's own forward process, and the same reverse process completes the
missing fine scales.

%% file: sections/3_prelim.tex
\section{Preliminaries and Motivations}\label{sec:prelims}

\textbf{Standard diffusion.}
Diffusion models \cite{ho2020denoising,song2020score} generate samples by
reversing a fixed forward noising process. The forward process gradually
transforms a data sample $\mathbf{x}_0$ into isotropic Gaussian noise via
\begin{equation}\label{eq:standard_ddpm}
    \mathbf{x}_t = \sqrt{\bar\alpha_t}\, \mathbf{x}_0
    + \sqrt{1-\bar\alpha_t}\, {\epsilon},
    \qquad {\epsilon} \sim \mathcal{N}(0, \mathbf{I}),
\end{equation}
where the schedule $\bar\alpha_t$ decreases monotonically from $1$ at $t=0$ to
nearly $0$ at the end of diffusion. Because the marginals are jointly Gaussian, the
reverse-time conditional $q(\mathbf{x}_{t-1} \mid \mathbf{x}_t, \mathbf{x}_0)$
is itself Gaussian and analytically tractable. A neural network
${\epsilon}_\theta(\mathbf{x}_t, t)$ is trained to predict
${\epsilon}$ given $\mathbf{x}_t$ by minimizing
$\mathbb{E}\big[\|{\epsilon} - {\epsilon}_\theta(\mathbf{x}_t, t)\|^2\big]$,
and substituting this prediction into the reverse posterior gives a tractable
sampling step. Iterative denoising starting from pure noise produces samples
from the data distribution.

\textbf{Scale invariance in physics.}
Critical physical systems, exemplified by the two-dimensional Ising model at
its critical temperature, exhibit scale invariance explicitly. Such systems have
no characteristic length scale, so configurations look statistically the same
after coordinates are coarse-grained and rescaled by any factor $\mathbf{r} \to b\mathbf{r}$. As
a consequence, statistical observables follow power laws of the form
$O(k) \propto k^{-\alpha}$ with universal exponents $\alpha$
\cite{onsager1944crystal,wilson1974renormalization,belavin1984infinite}, since
power laws are the only functions invariant under rescaling up to a
multiplicative constant. The renormalization group formalizes this picture.
Coarse-graining out fine-scale degrees of freedom and rescaling the
result acts as a transformation on probability distributions, and the
distribution of a critical system is a fixed point of that transformation.

\begin{figure}[t]
    \centering
    \begin{minipage}{0.55\textwidth}
        \centering
        \begin{subfigure}{0.48\linewidth}
            \centering
            \includegraphics[width=\linewidth]{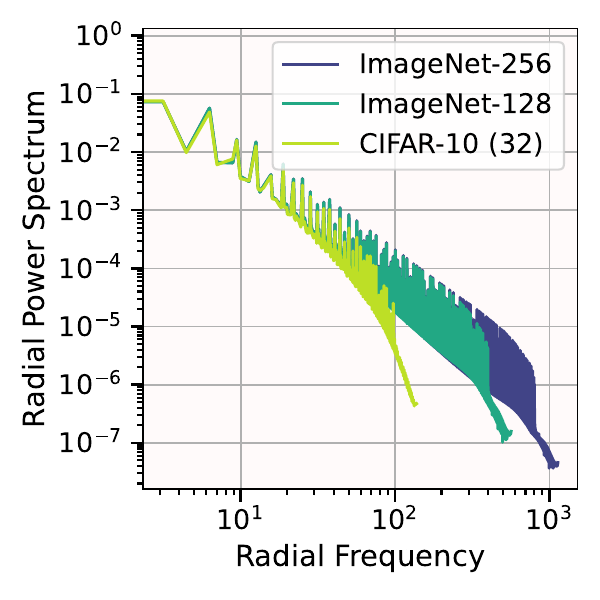}
            \caption{Natural-image variances}
            \label{fig:natural_variance}
        \end{subfigure}
        \hfill
        \begin{subfigure}{0.48\linewidth}
            \centering
            \includegraphics[width=\linewidth]{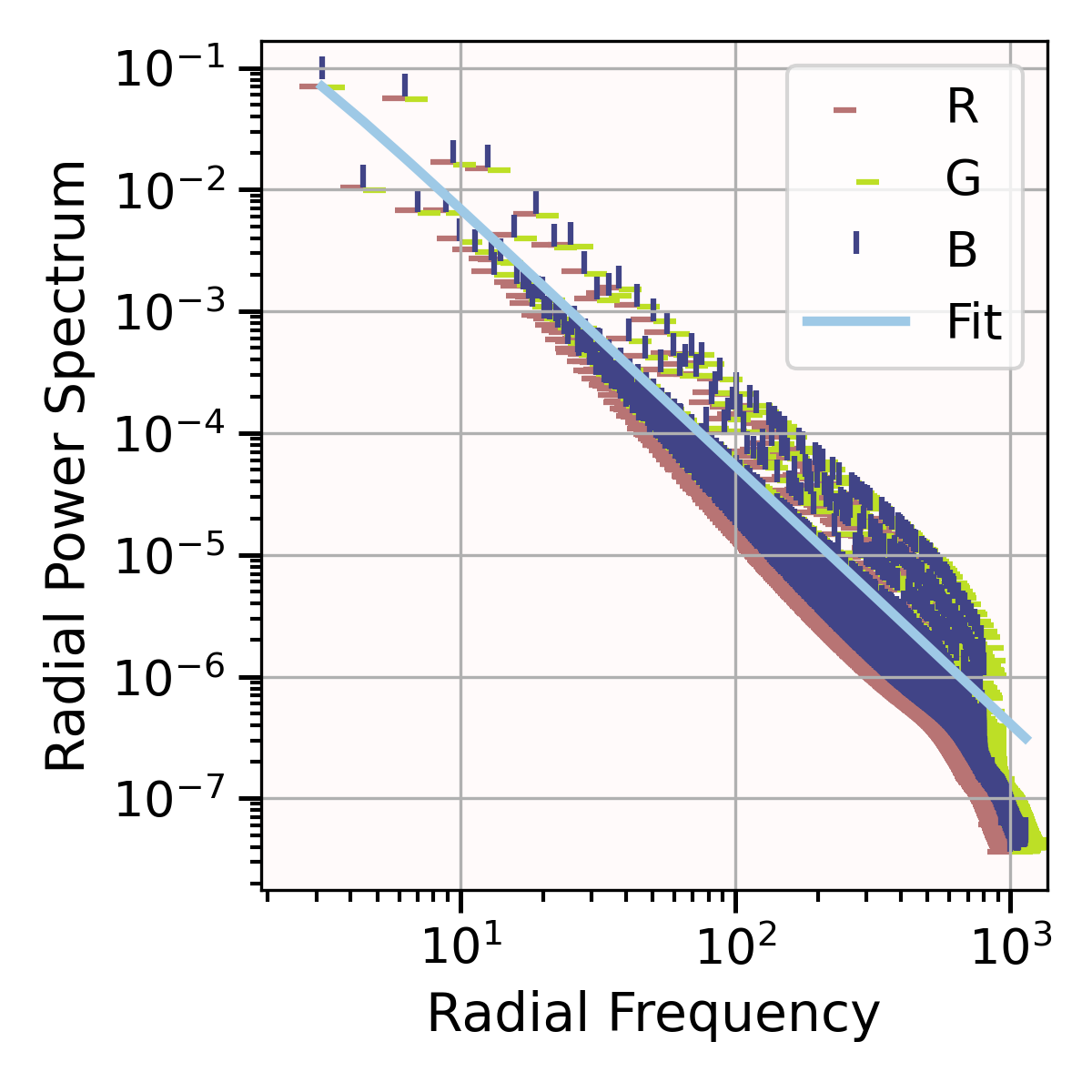}
            \caption{ImageNet-256 variances}
            \label{fig:imagenet256_variance}
        \end{subfigure}
    \end{minipage}%
    \hfill
    \begin{minipage}{0.4\textwidth}
        \captionsetup{justification=raggedright,singlelinecheck=false}
        \caption{Variance power spectra of natural-image datasets.
        (a) Spectra of CIFAR-10, ImageNet-128, and ImageNet-256 exhibit similar
        power-law decay over their shared frequency range, indicating
        approximate scale invariance.
        (b) Variance of ImageNet-256 computed independently for each color
        channel (RGB), with a power-law fit recovering the $k^{-2}$ frequency
        decay.}
        \label{fig:variance_summary}
    \end{minipage}
\end{figure}

\textbf{Power-law spectra of natural images.}
Natural-image distributions show approximate scale invariance. Their radially
averaged power spectra, equivalently the variance per Fourier mode of the
dataset $\mathbf{S}_0(\mathbf{k}) = \mathbb{E}\big[|{\mathbf{X}}_0(\mathbf{k})|^2\big] - \big |\mathbb{E}\big[{\mathbf{X}}_0(\mathbf{k})\big]\big |^2$,
closely follow $k^{-2}$ over a wide frequency range
\cite{field1987relations,ruderman1994statistics,vanderschaaf1996modelling,hyvrinen2009natural},
on average across a dataset. We confirm this on the datasets used in our
experiments. Figure~\ref{fig:variance_summary} shows the radially averaged
variance power spectra for CIFAR-10, ImageNet-128, and ImageNet-256 computed
in the discrete cosine transform (DCT) space \cite{ahmed1974discrete}, with the exact transform given in
Appendix~\ref{app:spectrum}. The spectra agree over their shared frequency
range and differ mainly near finite-resolution cutoffs. We fit the radial
variance with
\begin{equation}\label{eq:variance_fit}
    \mathbf{S}_0(\mathbf{k}) = C(\mathbf{k}^2 + \mathbf{k}_0^2)^{-a},
\end{equation}
where $\mathbf{k}_0$ regularizes the $\mathbf{k} \to \mathbf{0}$ limit. The
fits recover the $k^{-2}$ scaling, and
Table~\ref{tab:powerlaw_params} lists the fitted parameters.

\textbf{Toward scale-invariant diffusion.}
The observations above raise a natural design question. Given that natural
images and critical physical systems share a hierarchy of structure across
scales, what would a diffusion forward process look like if it were designed to respect this hierarchy rather than treating all scales on the same footing? We propose such a scale-invariant process in the frequency space in the next section.

%% file: sections/4_theory.tex
\section{Scale-Invariant Diffusion in Frequency Space}
\label{sec:method}

\subsection{Formulation}

\textbf{Forward process.}
Let $\mathbf{X}_0(\mathbf{k})$ denote the DCT coefficients of an image and
let $\mathbf{S}_0(\mathbf{k})$ be the empirical variance spectrum estimated
in Section~\ref{sec:prelims}. We define the continuous forward marginal
\begin{equation}
\label{eq:field-eq}
    \mathbf{X}(\mathbf{k}, t)
    =
    \underbrace{e^{-\mathbf{k}^2 \lambda(t)/2} \odot \mathbf{X}_0(\mathbf{k})}_{\text{signal}}
    +
    \underbrace{\sqrt{1 - e^{-\mathbf{k}^2 \lambda(t)}}
    \odot \sqrt{\mathbf{S}_0(\mathbf{k})}\, \epsilon_t}_{\text{noise}},
    \qquad
    \epsilon_t \sim \mathcal{N}(0, \mathbf{I}),
\end{equation}
where $\odot$ denotes Hadamard product.
The schedule $\lambda(t)$ is a scalar function that monotonically increases in $t$. As
$t$ grows, $e^{-\mathbf{k}^2 \lambda(t)/2}$ narrows in frequency space, so high-frequency modes are
attenuated before low-frequency ones. The noise prefactor is chosen so that the forward marginal preserves the per-mode covariance
$\mathbf{S}_0(\mathbf{k})$ in expectation at every $t$, and converges to
$\mathcal{N}(0, \mathbf{S}_0(\mathbf{k}))$ as the signal term vanishes. In
pixel space, the same process convolves the signal with a Gaussian kernel and adds spatially correlated noise
whose correlation length grows with $t$, reflecting the progressive removal
of scale structures.

\textbf{Discretization.}
All experiments use a DDPM \cite{ho2020denoising} discretization of Eq.~\eqref{eq:field-eq}. For $0 = t_0 < \cdots < t_N = 1$, let
$\bar{\boldsymbol{\alpha}}_n(\mathbf{k}) = e^{-\mathbf{k}^2 \lambda_n}$,
$\boldsymbol{\alpha}_n = \bar{\boldsymbol{\alpha}}_n /
\bar{\boldsymbol{\alpha}}_{n-1}$, and
$\boldsymbol{\beta}_n = 1 - \boldsymbol{\alpha}_n$. Then
\begin{equation}
\label{eq:kddpm}
    \mathbf{X}_n(\mathbf{k})
    =
    \sqrt{\bar{\boldsymbol{\alpha}}_n} \odot \mathbf{X}_0(\mathbf{k})
    +
    \sqrt{1 - \bar{\boldsymbol{\alpha}}_n}
    \odot \sqrt{\mathbf{S}_0(\mathbf{k})}\, \epsilon_n,
    \qquad
    \epsilon_n \sim \mathcal{N}(0, \mathbf{I}).
\end{equation}
The one-step transition has the same form with
$\bar{\boldsymbol{\alpha}}_n$ replaced by $\boldsymbol{\alpha}_n$.

Since all covariances are diagonal in frequency space, the reverse
posterior is Gaussian:
$q(\mathbf{X}_{n-1} \mid \mathbf{X}_n, \mathbf{X}_0)
= \mathcal{N}(\boldsymbol{\mu}_q, \mathbf{S}_0 \tilde{\boldsymbol{\beta}}_n)$,
with
\begin{equation}
\label{eq:posterior-coeffs}
    \tilde{\boldsymbol{\beta}}_n
    = \frac{\boldsymbol{\beta}_n (1 - \bar{\boldsymbol{\alpha}}_{n-1})}
           {1 - \bar{\boldsymbol{\alpha}}_n},
    \qquad
    \boldsymbol{\mu}_q(n, \mathbf{X}_n)
    = \frac{1}{\sqrt{\boldsymbol{\alpha}_n}}
      \left( \mathbf{X}_n
      - \frac{\boldsymbol{\beta}_n \sqrt{\mathbf{S}_0}}
             {\sqrt{1 - \bar{\boldsymbol{\alpha}}_n}}\,
             \epsilon_n \right).
\end{equation}
Ancestral sampling proceeds by
\begin{equation}
\label{eq:ddpm-sample}
    \mathbf{X}_{n-1}
    = \boldsymbol{\mu}_q(n, \mathbf{X}_n)
    + \sqrt{\mathbf{S}_0 \tilde{\boldsymbol{\beta}}_n} \odot \epsilon_n.
\end{equation}
The full DDPM and stochastic differential equation (SDE) derivation appear in
Appendices~\ref{app:ddpm-theory} and~\ref{app:sde-theory}.

\subsection{Training target and numerical cutoffs}

We train an $\epsilon$-prediction network with the loss
\begin{equation}
    \mathcal{L}
    =
    \mathbb{E}_{n, \mathbf{X}_0, \epsilon}
    \bigl[
        \lVert \epsilon - \epsilon_\theta(n, \mathbf{X}_n) \rVert_2^2
    \bigr].
\end{equation}
Two implementation details make the finite-resolution process stable.
First, very small $\boldsymbol{\alpha}_n(\mathbf{k})$ values can cause large
reverse updates for high-frequency modes, so we floor them at $10^{-6}$ in
the ancestral sampler. Second, the zero mode would otherwise have no
attenuation or noise because the Gaussian signal filter leaves it untouched. Therefore, we introduce a
low-frequency cutoff $k_c$, and use $\max(\lVert\mathbf{k}\rVert, k_c)$ as the schedule for modes with
$\lVert\mathbf{k}\rVert \leq k_c$. This preserves the algebra above
while properly handles the low-frequency limit, where scale-invariance is affected by finite size effect.

\subsection{Schedules and effective resolution}

The schedule $\lambda(t)$ controls how frequencies are attenuated with time. We evaluate two schedules, named by how the damping cutoff in
$\mathbf{k}$ moves with time. The log-linear schedule $\lambda(t) = t \cdot 10^{\lambda_i + (\lambda_f - \lambda_i)\, t}$ moves it roughly uniformly
on a log scale, and the linear schedule $\lambda(t) = \theta \, t/(\lambda_f (1 - t) + \lambda_i)^2$ moves it roughly uniformly on a linear scale. The multiplicative $t$ ensures $\lambda(0) = 0$. Among the parameters,
$\lambda_i$ primarily sets the high-frequency, early-time behavior, while
$\lambda_f$, $k_c$, and $\theta$ primarily set the low-frequency,
late-time behavior; all four jointly shape the full schedule.

The notion of time-evolving cutoff in $\mathbf{k}$ gives super-resolution a direct interpretation. For a chosen
SNR threshold,
\begin{equation}
    \mathrm{SNR}_n(\mathbf{k})
    = \frac{\bar{\boldsymbol{\alpha}}_n(\mathbf{k})}
           {1 - \bar{\boldsymbol{\alpha}}_n(\mathbf{k})},
\end{equation}
the modes above the threshold define an effective resolution. Starting the
reverse process from a timestep where the effective resolution is zero
gives image generation; starting from a timestep whose surviving signals correspond to a lower-resolution input gives super-resolution.
Because $\lambda(t)$ is continuous before discretization and can be densely
sampled in implementation, the effective resolution varies continuously along the schedule, yielding a continuum of super-resolution
factors from a single trained model (Figure~\ref{fig:res_sched}).

\subsection{Connection to scale-space theory and renormalization group}

Equation~\eqref{eq:field-eq} extends the vanilla scale-space operation \cite{koenderink1984structure, lindeberg1994scale} to frequency space
with noise. In pixel space, multiplying DCT modes by
$\exp[-\mathbf{k}^2 \lambda(t) / 2]$ amounts to Gaussian smoothing at scale
$\lambda(t)$, the same operator that appears in linear scale-space theory. Crucially, our additional noise
term turns the deterministic smoothing into a stochastic coarse-graining
process whose final covariance matches the dataset variance spectrum.

The same equation also conceptually resembles a renormalization group (RG)
coarse-graining step, where short-distance degrees of freedom are discarded before long-distance structures. We do not claim that our method
is an exact RG transformation; rather, we use the RG as analogy:
if a dataset exhibits approximate scale invariance, a reverse
model trained on the scale-ordered forward process should learn how fine
scales are distributed conditioned on coarse scales. The critical-Ising
experiment in Section~\ref{sec:scientific-data} tests this idea in a setting
with known scale-invariant structure.

%% file: sections/5_experiment.tex
\section{Experiments}

We evaluate whether the same frequency-space diffusion process can serve as an
unconditional image generator and a continuous super-resolution model. We test
SKILD in three settings: unconditional image generation on CIFAR-10,
$2\times$–$8\times$ continuous super-resolution on ImageNet-$128$ and $256$, and
a scientific benchmark on the critical two-dimensional Ising model, probing
scale invariance directly through four-point correlations. SKILD is competitive
with or outperforms strong baselines in each setting.

\subsection{Setup}
\textbf{Architecture.}
All reported models use a score U-Net backbone from the NCSN++ family
\cite{song2020score}. Exact channel counts, depths, and attention configurations
are in Appendices~\ref{app:cifar} and~\ref{app:imagenet}.

\textbf{Data.}
We use CIFAR-10 \cite{krizhevsky2009learning} and ImageNet
\cite{deng2009imagenet} as released. Critical-Ising configurations are sampled
on a $128 \times 128$ square lattice using the Wolff cluster algorithm
\cite{wolff1989collective}; data-generation details are in
Appendix~\ref{app:ising-sr}.

\textbf{Noise schedule.}
We test both the log-linear and linear schedules on CIFAR-10 and the linear
schedule on ImageNet and Ising experiments. All experiments use $N=1000$
timesteps; exact schedule parameters are in
Appendices~\ref{app:cifar},~\ref{app:imagenet}, and~\ref{app:ising-sr}.

\textbf{Training.}
We train with AdamW and use an exponential moving average of weights at sampling
time. Full hyperparameters and compute details are in
Appendices~\ref{app:cifar},~\ref{app:imagenet}, and~\ref{app:ising-sr}.

\textbf{Evaluation.}
For CIFAR-10 we report FID \cite{heusel2017gans} and Inception Score (IS)
\cite{salimans2016improved} on $50$K generated samples. For ImageNet
super-resolution we report PSNR, SSIM \cite{wang2004image}, LPIPS
\cite{zhang2018unreasonable}, MUSIQ \cite{ke2021musiq}, and CLIPIQA
\cite{wang2023exploring} from the last checkpoint, on a random $3$K-image subset
of the ImageNet validation set, following the protocols of
\cite{yue2023resshift,wang2024sinsr}. The Ising super-resolution experiment is
evaluated on $1$K samples from the last checkpoint by a connected four-point
correlation, a statistical-physics observable that probes how accurately the
model reproduces Ising model structures across scales
(Section~\ref{sec:scientific-data}).

\subsection{Effective resolution protocol}

For super-resolution, we choose the reverse starting timestep by the SNR defined
in Section~\ref{sec:method}. We use threshold $0.1$ in all ImageNet and Ising
super-resolution experiments. This corresponds to applying an effective low-pass
filter in the frequency domain, where modes below the effective-resolution
cutoff retain signal while higher-frequency modes are attenuated and dominated
by noise. At sampling time, the model starts from the exact forward marginal of
the paired high-resolution (HR) image at the chosen timestep, then reverses to
$t_0$. This protocol turns super-resolution into a partial reverse diffusion
problem rather than a conditional generation problem.

We validate the effective-resolution interpretation by comparing the surviving
signal at the chosen timestep with standard bicubic down-up sampling. The MSE
and PSNR values in Table~\ref{tab:gt_bicubic} show that SNR of $0.1$ produces
low-resolution (LR) inputs close to conventional $4\times$ or $8\times$
degradations while preserving consistency with the forward diffusion process.

\subsection{Unconditional CIFAR-10 generation}

\newcolumntype{C}[1]{>{\centering\arraybackslash}m{#1}}

\begin{table}[ht!]
    \centering
    \footnotesize
    \setlength{\tabcolsep}{1.5pt}
    \caption{Unconditional CIFAR-10 sample quality. Best results within each
    model group are in \textbf{bold}.}
    \label{tab:cifar10}
    \begin{threeparttable}
    \begin{tabular}{m{.06\linewidth}| C{.06\linewidth} 
    C{.08\linewidth} 
    C{.08\linewidth} 
    C{.06\linewidth} 
    | C{.1\linewidth} 
    C{.1\linewidth} 
    C{.08\linewidth} 
    C{.08\linewidth} 
    C{.06\linewidth} 
    C{.07\linewidth} 
    C{.06\linewidth} 
    }
    \toprule
    \multirow{2}{*}{Model}
    & \multicolumn{4}{c|}{\textbf{Frequency-agnostic model}} &
    \multicolumn{7}{c}{\textbf{Frequency/scale-informed model}} \\
    & DDPM \cite{ho2020denoising} & DDPM++ \cite{song2020score} & NCSN++
    \cite{song2020score} & EDM \cite{karras2022elucidating} & Cold diff.*
    (deblur) \cite{bansal2023cold} & Cold diff.* (sr) \cite{bansal2023cold} &
    EqualSNR \cite{falck2025fourier} & WaveDiff \cite{guth2022wavelet} & IHDM
    \cite{rissanen2022generative} & DCTdiff \cite{ning2024dctdiff} & SKILD
    (ours) \\
    \midrule
    FID $\downarrow$ & 3.17 & 2.78 & 2.20 & \textbf{1.97} & 80.08 & 152.76&
    13.63 & 4.01 & 18.96 & 5.02  & \textbf{2.61} \\
    IS $\uparrow$ & -- & 9.46 & 9.64 & \textbf{9.89} & -- & -- & -- & -- & -- &
    7.70 & \textbf{9.65} \\
    \bottomrule
    \end{tabular}
    \begin{tablenotes}[flushleft]
        \footnotesize
        \item[*] The \cite{bansal2023cold} results are generated from
        highly-degraded images which still contains some signal, instead from
        scratch.
    \end{tablenotes}
    \end{threeparttable}
\end{table}

\begin{figure}[ht!]
    \centering
    \includegraphics[width=1.0\linewidth]{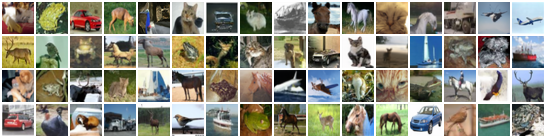}
    \caption{Uncurated samples of generated images on CIFAR-10; more in
    Appendix~\ref{app:img_sample} (Figure~\ref{fig:cifar_600}).}
    \label{fig:cifar_64}
\end{figure}

\begin{figure}[ht!]
    \centering
    \begin{minipage}{0.6\textwidth}
        \centering
        \includegraphics[width=\linewidth]{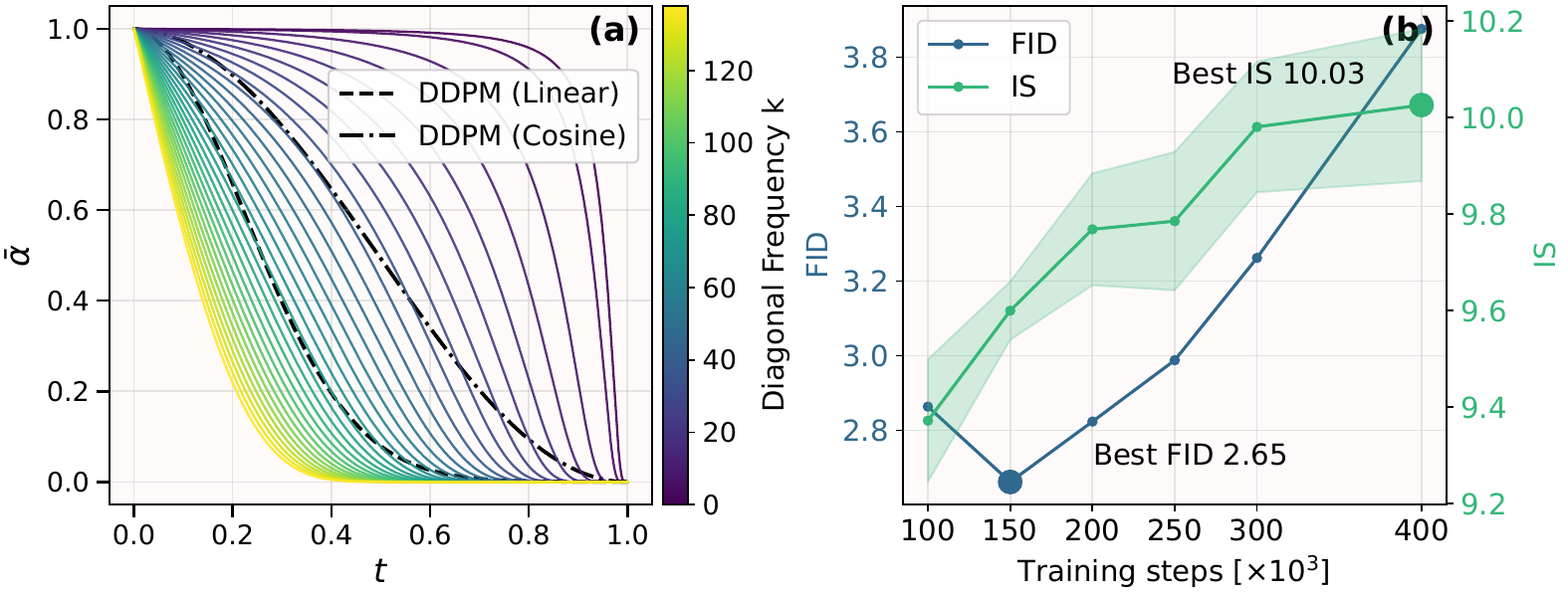}
    \end{minipage}%
    \hfill
    \begin{minipage}{0.38\textwidth}
        \captionsetup{justification=raggedright,singlelinecheck=false}
        \caption{(a) Linear schedule with best FID and IS on CIFAR-10 generation. Lower-frequency modes are attenuated later than higher-frequency ones. (b) Mode collapse with training steps. While IS continues to improve with training, FID converges early on and worsens at later times.}
        \label{fig:sched_ablation}
    \end{minipage}
\end{figure}

Table~\ref{tab:cifar10} compares unconditional CIFAR-10 generation.
Figure~\ref{fig:cifar_64} shows uncurated samples that SKILD generates on
CIFAR-10. SKILD is competitive with the state-of-the-art models and achieves the
best FID and IS among the frequency or scale-informed models listed, using the
linear schedule shown in Figure~\ref{fig:sched_ablation}(a).

\textbf{Ablation studies.}
We conduct several sets of ablation studies and discuss three of them
briefly here. Figure~\ref{fig:sched_ablation}(b) shows the model over training.
A mode collapse appears with training course: FID reaches its best value before
the final checkpoint, while IS continues to improve. We interpret this as
evidence that low-frequency reconstruction remains the bottleneck for generation
on object-centric datasets like CIFAR-10. The limitation section discusses this
point directly.

\begin{figure}[ht!]
    \centering
    \includegraphics[width=0.75\linewidth]{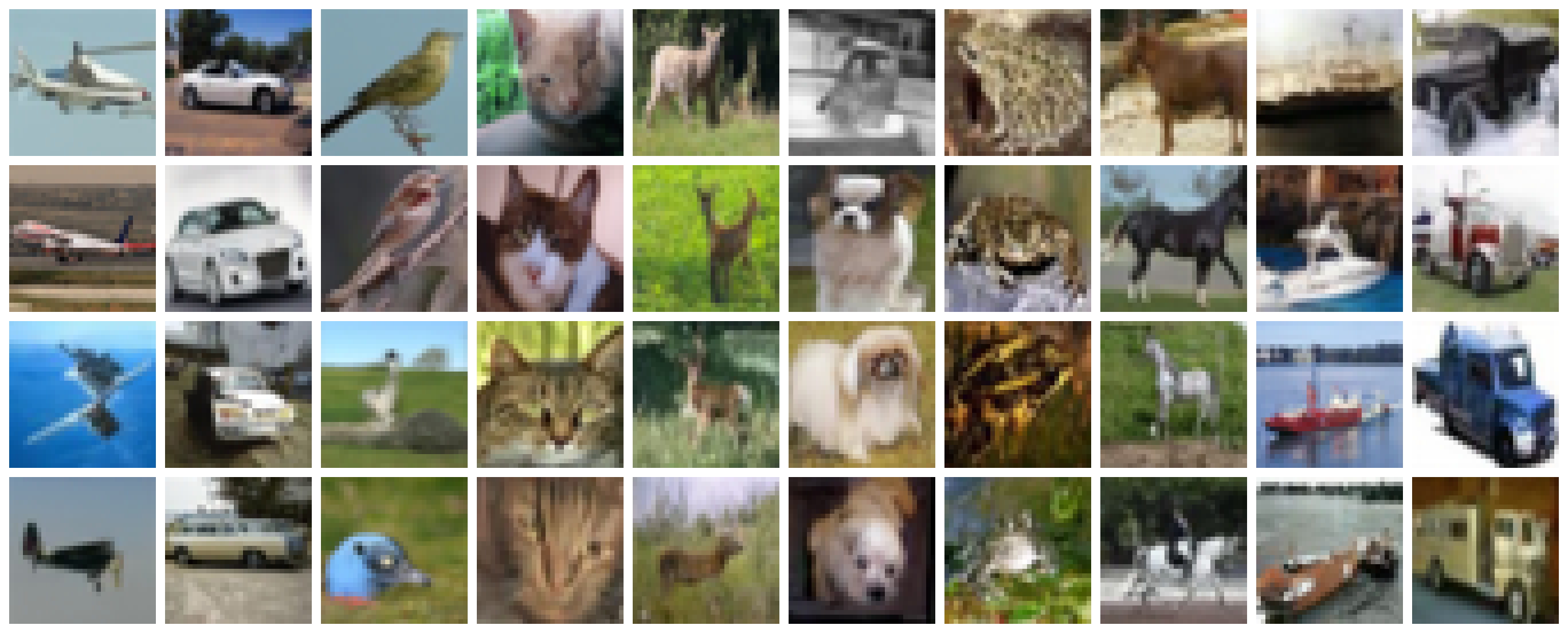}
    \caption{Class-conditional CIFAR-10 samples generated by SKILD with
    classifier-free guidance ($s = 1.2$). From left to right, each column shows
    airplane, automobile, bird, cat, deer, dog, frog, horse, ship, and truck.}
    \label{fig:class_cond}
\end{figure}

This mode collapse is not inherent to SKILD. It happens at the end of the
forward process, where only the lowest frequencies still carry signal
(Appendix~\ref{app:mode-collapse}). Scale invariance breaks down at the lowest
frequencies for images of finite size and resolution, much like finite system
size and lattice spacing break scale invariance in critical physical systems.
This final stage makes up only a small fraction of the diffusion process and
needs separate handling. We test two fixes using the same U-Net, optimizer,
dataset, training budget, and schedule (Table~\ref{tab:lowfreq}). The first
option pretrains a small $19$M-parameter VAE \cite{kingma2014auto} on the
heavily blurred images from the final diffusion steps, reaching a reconstruction
mean squared error of $2.35 \times 10^{-7}$. During sampling, the frozen VAE
generates this initial low-frequency state and the reverse process runs from
there, keeping the whole pipeline unconditional. The second option adds class
labels and uses classifier-free guidance \cite{ho2022classifier} with a scale of
$s = 1 + w = 1.2$, where $w$ is the guidance weight. Both methods keep training
stable through $400$K steps, and both FID and IS improve between $200$K and
$400$K steps. Figure~\ref{fig:class_cond} shows samples from the
class-conditional model. Because the guided model does not use the VAE, the two
fixes are independent and can be used together to further improve the
performance of model.

\begin{table}[ht!]
    \centering
    \small
    \caption{Two solutions for low-frequency mode collapse on CIFAR-10. Both
    runs share the same training setup and model configurations. The network used has half the channel count compared to the one in Table \ref{tab:cifar10} and Figure \ref{fig:sched_ablation}(b).}
    \vspace{0.1 in}
    \label{tab:lowfreq}
    \begin{tabular}{lcccc}
    \toprule
    & \multicolumn{2}{c}{FID $\downarrow$} & \multicolumn{2}{c}{IS $\uparrow$}
    \\
    \cmidrule(lr){2-3}\cmidrule(lr){4-5} Add-ons & 200K & 400K & 200K & 400K \\
    \midrule
    Low-frequency VAE & 3.76 & 3.33 & 9.41 & 9.63 \\
    Class conditioning + CFG ($s = 1.2$) & 3.43 & 3.33 & 9.48 & 9.75 \\
    \bottomrule
    \end{tabular}
\end{table}

\vspace{-0.1 in}

In Table~\ref{tab:schedule-robustness}, we verify the robustness of SKILD on
image generation against a broad range of schedules in the log-linear and linear
families. Most schedules reach FID below or near $5$ and all reach IS near or
above $9$ within $400$K training steps. This indicates that SKILD is not tuned
to a single fragile schedule, albeit the convergence speed is
schedule-dependent.

SKILD combines spectrum-matched noise with frequency-ordered attenuation to achieve scale invariance.
In Table~\ref{tab:mechanism}, we remove each component one at a time while
keeping the rest of the model configuration and training fixed. First, we
replace $\mathbf{S}_0$ with a nearly flat spectrum, which resembles
standard white noise rather than the $k^{-2}$ power law of natural images. This
change pushes FID at $400$K steps up to $20.02$. This drastic degradation shows the importance
of a spectrum-matched noise, which is necessary to maintain scale invariance in the diffusion process
\footnote{$\mathbf{S}_0$ is cheap to compute -- estimating $\mathbf{S}_0$ on ImageNet
takes only a few minutes on a single consumer GPU. It is also similar across natural image datasets.}. Second, we keep the true
$\mathbf{S}_0$ but replace the frequency-dependent attenuation $\bar{\boldsymbol{\alpha}}_n(\mathbf{k})$ with the
standard DDPM schedule \cite{ho2020denoising}, which attenuates all
frequencies at the same rate. This also hurts both FID and IS significantly.
The ablation demonstrates that both components are necessary to ensure scale invariance
and achieve high-quality image generation with SKILD.

\begin{table}[ht!]
    \centering
    \small
    \caption{Ablations on the effect of spectrum-matched noise and frequency-dependent attenuation in SKILD.
    Without either component, FID and IS noticeably degrade.
    }
    \vspace{0.1 in}
    \label{tab:mechanism}
    \begin{tabular}{lcccc}
    \toprule
    & \multicolumn{2}{c}{FID $\downarrow$} & \multicolumn{2}{c}{IS $\uparrow$}
    \\
    \cmidrule(lr){2-3}\cmidrule(lr){4-5} Ablations & 200K & 400K & 200K & 400K \\
    \midrule
    flat $\mathbf{S}_0$ & 18.23 & 20.02 & 8.14 & 7.96 \\
    Frequency-agnostic $\bar{\boldsymbol{\alpha}}_n$ & 6.59 & 5.35 & 8.66 &
    8.93 \\
    \bottomrule
    \end{tabular}
\end{table}

More details of ablations can be found in Appendix~\ref{app:ablation}, including
FID and IS convergence over training compared to common pixel-space diffusion
schedules (Figure~\ref{fig:cifar-train}), different network predictions, the
effectiveness of second-moment sampler, potential of reducing number of
diffusion steps, and different numerical cutoffs.


\subsection{ImageNet super-resolution}

\begin{table*}[ht!]
    \centering
    \small
    \caption{$4\times$ super-resolution quality on the ImageNet-Test
    \cite{yue2023resshift,wang2024sinsr}. Best and second-best results among
    quality metrics are highlighted in \textbf{bold} and \underline{underline}.}
    \label{tab:sr_metrics}
    \begin{tabular}{lccccc|c}
    \toprule
    Model & PSNR$\uparrow$ & SSIM$\uparrow$ & LPIPS$\downarrow$ &
    CLIPIQA$\uparrow$ & MUSIQ$\uparrow$ & \# Param. (M) \\
    \midrule
    \multicolumn{6}{l|}{\textit{GAN-based}} & \\
    \midrule
    BSRGAN~\cite{zhang2021designing} & 24.42 & 0.659 & 0.259 & 0.581 &
    \underline{54.697} & 16.70 \\
    SwinIR~\cite{liang2021swinir} & 23.99 & 0.667 & 0.238 & 0.564 & 53.790 &
    16.70 \\
    RealESRGAN~\cite{wang2021real} & 24.04 & 0.665 & 0.254 & 0.523 & 52.538 &
    28.01 \\
    \midrule
    \multicolumn{6}{l|}{\textit{Conditional diffusion-based}} & \\
    \midrule
    LDM-30~\cite{rombach2022high} & 24.49 & 0.651 & 0.248 & 0.572 & 50.895 &
    113.60 \\
    LDM-15~\cite{rombach2022high} & 24.89 & 0.670 & 0.269 & 0.512 & 46.419 &
    113.60 \\
    ResShift~\cite{yue2023resshift} & 25.01 & 0.677 & 0.231 & 0.592 & 53.660 &
    118.59 \\
    SinSR~\cite{wang2024sinsr} & 24.56 & 0.657 & 0.221 & \underline{0.611} &
    53.357 & 118.59 \\
    IRSDE~\cite{luo2023image} & 24.48 & 0.602 & 0.304 & 0.513 & 45.382 & 137.20
    \\
    DDRM~\cite{kawar2022denoising} & \underline{25.56} & 0.674 & 0.471 & 0.372 &
    24.746 & 552.80 \\
    I2SB~\cite{liu20232} & \textbf{26.76} & \textbf{0.730} & \underline{0.206} &
    0.489 & 53.936 & 552.80 \\
    \midrule
    \multicolumn{6}{l|}{\textit{Unconditional diffusion}} & \\
    \midrule
    {SKILD} (Ours) & 24.10 & \underline{0.683} & \textbf{0.186} & \textbf{0.612}
    & \textbf{59.226} & 121.12 \\
    \bottomrule
    \end{tabular}
\end{table*}

\begin{figure}[ht!]
    \centering
    \includegraphics[width=.75\linewidth]{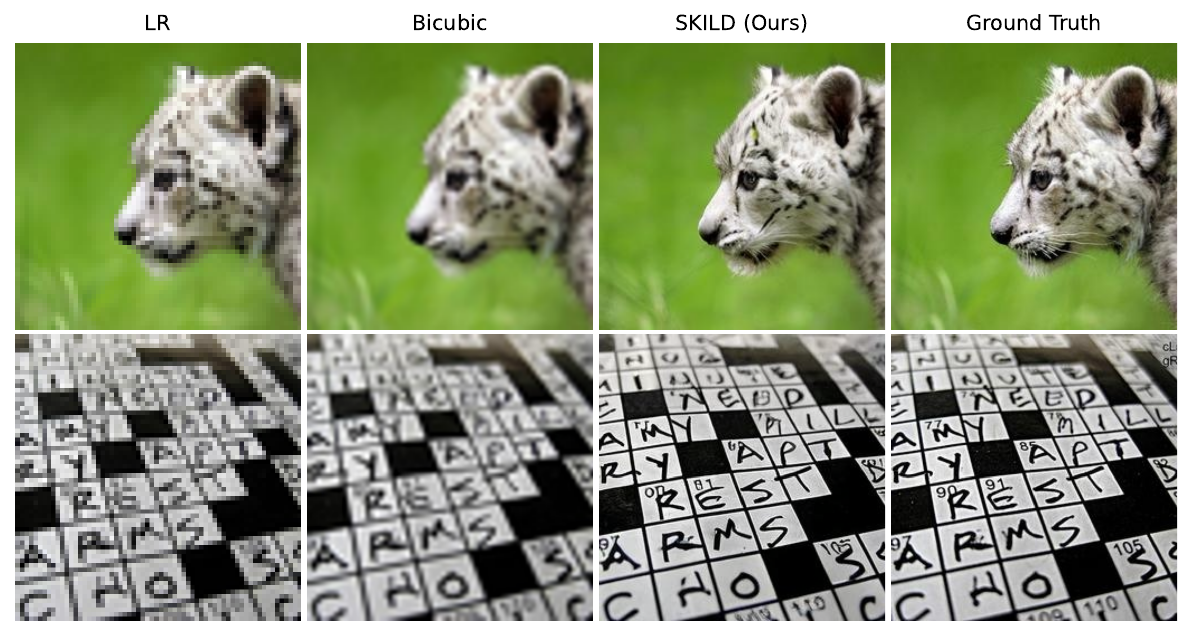}
    \caption{$4\times$ super-resolution samples on ImageNet-256. The model is
    initialized from a $64\times64$ low-resolution forward state and
    reconstructs high-frequency details through the reverse process. }
    \label{fig:256_64_2}
\end{figure}

\begin{figure}[ht!]
    \centering
    \includegraphics[width=1.0\linewidth]{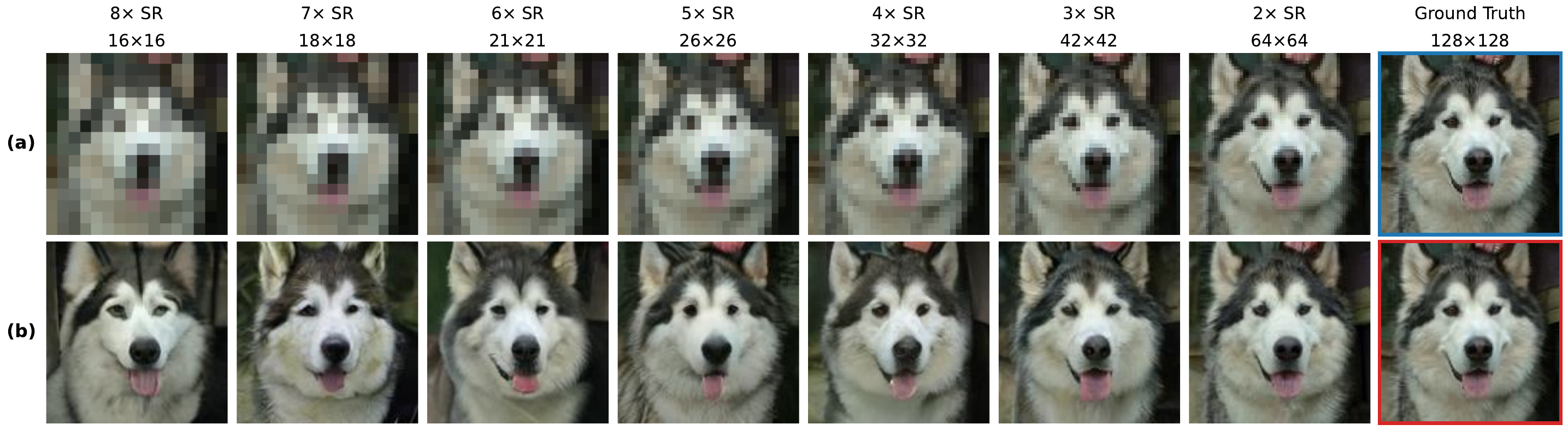}
    \caption{Continuous super-resolution on ImageNet-128. (a) Low-resolution
    inputs at effective resolutions $16\times16$ through $64\times64$,
    corresponding to $8\times$ through $2\times$ super-resolution factors. (b)
    High-resolution reconstructions produced by the same checkpoint from each
    effective-resolution starting state. A continuum of super-resolution factors
    is accessible from a single trained model.}
    \label{fig:cont-sr}
\end{figure}

Table~\ref{tab:sr_metrics} reports $4\times$ super-resolution quality on
ImageNet. All conditional baselines receive the low-resolution input through an
explicit conditioning path, most of which additionally use class labels or
classifier-free guidance. SKILD uses no conditioning of any kind: it starts from
the corresponding forward marginal and runs the same reverse process used for
unconditional generation. Despite this simplicity in design, the
$256$-resolution model achieves the best LPIPS, CLIPIQA, and MUSIQ among the
compared methods and the second-best SSIM. PSNR favors methods with higher raw
pixel accuracy, while metrics that emphasize human perception favor our method.
This gap comes from the perception-distortion tradeoff
\cite{blau2018perception}. Minimizing mean squared error pushes a model to
predict the conditional mean, which averages over all the sharp images that
could match a given low-resolution input. Models trained this way get higher
PSNR by blurring out fine details. Instead of predicting an average, SKILD
samples the missing high frequencies through the reverse process, trading raw
pixel accuracy for sharper perceptual quality. Two super-resolution samples are
shown in Figure \ref{fig:256_64_2}, with more in Appendix \ref{app:img_sample}.

A single trained ImageNet model accommodates a continuum of super-resolution
factors by varying only the starting timestep. Figure~\ref{fig:cont-sr} shows
reconstructions at factors from $2\times$ to $8\times$ produced by the same
checkpoint, and Figure~\ref{fig:res_sched} plots how the effective resolution
decreases continuously with $t$.

%% file: sections/6_scientific_data.tex
\subsection{Scientific benchmark}
\label{sec:scientific-data}

\begin{figure}[ht!]
    \centering
    \includegraphics[width=\linewidth]{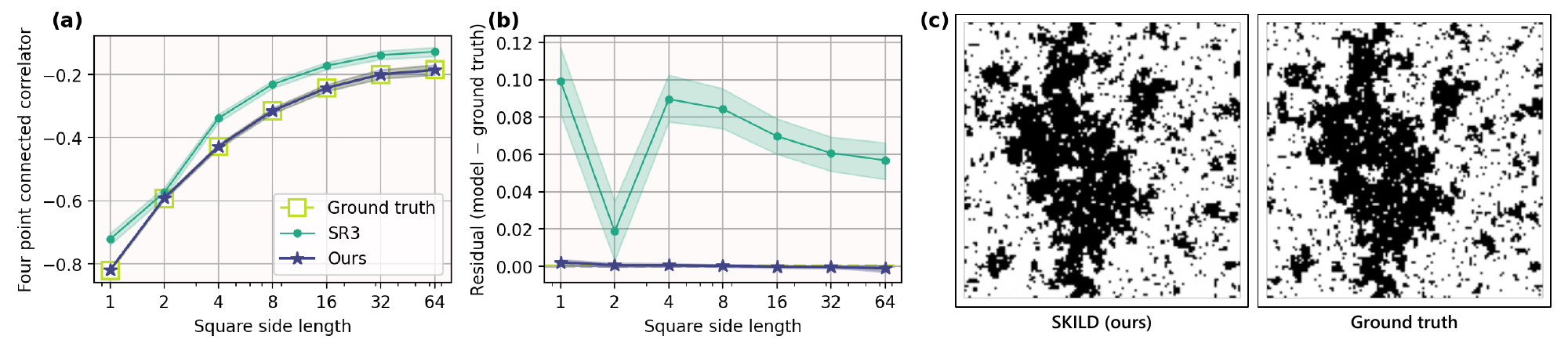}
    \caption{
    (a-b) Benchmark of four-point correlator accuracy. Our
    reconstruction closely tracks the ground truth, while SR3 shows a clear deviation.
    (c) SKILD-reconstructed critical Ising field sample compared to the ground truth.
    }
    \label{fig:ising-four-point}
\end{figure}

Natural images are approximately scale-invariant only after
averaging over many scenes. Critical physical systems let us ask a
stricter question: can a model reconstruct missing fine scales while
preserving observables that define the scale-invariant law? We test this
on the prototypical two-dimensional Ising model at criticality. Despite its simplicity--placing a spin $s_i \in \{-1,+1\}$ on each lattice site, with nearest
neighbors preferring to align--the Ising model serves a foundational role in areas across statistical mechanics, combinatorics, and computational complexity theory. At its critical temperature, the
correlation length diverges, the distribution becomes statistically
self-similar under RG coarse-graining, and the continuum limit is
described by a conformal field theory \cite{onsager1944crystal,
wilson1974renormalization,belavin1984infinite}. Previous works also have applied neural networks
to Ising super-resolution and inverse RG \cite{efthymiou2019super,
shiina2021inverse}.

This setting provides a more precise benchmark of scale-invariance than perceptual realism. A
visually plausible spin configuration can still have the wrong connected
correlations, the wrong response functions, or the wrong
universality-class signatures. 

\textbf{Evaluation: connected four-point correlator.}
We evaluate a connected four-point correlator, equivalently a
fourth-order joint cumulant, over the four corner spins of square
patches at multiple side lengths. The join cumulant subtracts all pairwise
contributions, isolating non-Gaussian dependence that cannot be inferred
from the mean, variance, or two-point correlation alone. Higher-order
correlations are central observables in critical systems, so matching
them across scales is a stronger test than matching visual texture or
pixel-level distortion. Data generation, paired evaluation, and the
correlator estimation are detailed in Appendix~\ref{app:ising-sr}.

\textbf{Results.}
We super-resolve from a $32\times32$ effective-resolution starting state
to a $128\times128$ critical-Ising field.
Figure~\ref{fig:ising-four-point}  shows that SKILD's reconstructed
four-point correlator closely tracks the ground truth at
every patch size, while SR3 \cite{saharia2022image}, a strong diffusion-based conditional super-resolution model, deviates significantly from the ground truth.

%% file: sections/7_conclusion.tex
\section{Conclusions}

We introduce SKILD, a scale-invariant frequency-space diffusion model whose forward
process is a stochastic coarse-graining operator: fine modes are damped
before coarse ones, and the injected noise carries the dataset spectrum.
Aligning diffusion with the scale structure of data makes scale an
explicit coordinate of the generative process. Unconditional generation
and super-resolution then become different starting points of the same
reverse trajectory, eliminating task-specific conditioning and guidance. Empirically,
SKILD reaches FID $2.61$ and IS $9.65$ on unconditional CIFAR-10,
supports continuous $2\times$--$8\times$ super-resolution from a single
ImageNet checkpoint, and reconstructs critical-Ising fields whose
connected four-point correlations closes tracks the ground truth.

\textbf{Discussion.}
Our method shifts the modeling burden from conditional mappings to
the design of a scale-invariant diffusion process. Rather than learning a
separate low-to-high-resolution mapping, the model learns a single
reverse trajectory over scales, with low-resolution inputs as
intermediate forward states. This removes task-specific conditioning or guidance. In the mean time, low-frequency generation becomes a central bottleneck: errors in coarse
structure early in the reverse chain propagate and constrain later
high-frequency generation. Our results reflect this trade-off:
super-resolution benefits from accurate coarse initialization, while
unconditional generation remains sensitive to low-frequency modeling.
The framework also suggests an evaluation criterion for self-similarity beyond perceptual
quality, namely whether fine-scale details remain statistically similar with coarse-scale structure, which is particularly relevant for scientific
applications. We provide such an instance through super-resolution experiments on critical-Ising fields.

\textbf{Limitations and future work.}
The current model opens a vast range of directions for future work.
(i) Currently, sampling requires $1000$ ancestral steps; faster samplers or
tailored solvers for mode-dependent schedules are a natural next step.
(ii) Unconditional generation remains sensitive to low-frequency
structure generation, and improvements there should stabilize global structure
without sacrificing fine detail.
(iii) Our super-resolution protocol uses exact forward marginals as
low-resolution inputs; extending to real-world degradations such as
unknown camera and compression pipelines is an important direction.
(iv) The Ising experiment establishes scale-invariant diffusion on one
critical system; extending to additional physical systems and
higher-order observables would broaden the scientific benchmark. (v) Our model uses off-the-shelf neural network. New architecture designs tailored to our model could significantly improve its performance while furthering understanding of network designs for diffusion models.

\textbf{Broader impacts.}
Unifying generation and super-resolution into a single reverse process
can simplify deployment pipelines and reduce reliance on task-specific
models. The same capability has dual-use risks: super-resolution can
hallucinate plausible but incorrect high-frequency content, with
consequences in forensics, medical imaging, and other sensitive settings
where outputs require validation. The scientific-data setting
illustrates a complementary benefit: scale-invariant models can be evaluated
against known physical laws, a style of evaluation we expect to be
useful wherever multi-scale structure carries scientific meaning.

\textbf{Code and reproducibility.}
The code for reproducing the main results and data of this paper is available at \texttt{https://github.com/JazzyCH/SKILD}.


\section*{Acknowledgment}

The authors acknowledge support from the National
Science Foundation under Cooperative Agreement PHY-2019786
(\href{http://iaifi.org/}{The NSF AI Institute for Artificial Intelligence and
Fundamental Interactions}) and the \href{https://genai.mit.edu/}{MIT Generative
AI Impact Consortium}. This work is supported by the Toyota Research Institute University 3.0 Program, and the Department of the Air Force Artificial Intelligence Accelerator under Cooperative Agreement No. FA8750-19-2-1000.
ZJC is in part supported by the Kurt Forrest Foundation Fellowship and the Henry Kendall Fellowship. ZC is in part supported by the MathWorks Fellowship and the Henry Kendall Fellowship. AW is in part supported by the National Science Foundation Graduate Research Fellowship. CD is in part supported by the Tayebati Postdoctoral Fellowship.
The authors also acknowledge the National Artificial Intelligence Research Resource (NAIRR) Pilot Program, the DeltaAI advanced computing and data resource at the National Center for Supercomputing Applications (supported by NSF Award OAC-2320345 and the State of Illinois), and Lambda Inc. for providing compute resources.

%% file: sections/10_appendix.tex
\newpage
\appendix

\renewcommand{\thefigure}{\thesection.\arabic{figure}}
\renewcommand{\thetable}{\thesection.\arabic{table}}
\renewcommand{\theequation}{\thesection.\arabic{equation}}
\renewcommand{\thesubsection}{\thesection.\Roman{subsection}}

\input{sections/10_1_app_theory}

\input{sections/10_2_app_exp}

%% file: sections/10_1_app_theory.tex
\section{DDPM formulation}\label{app:ddpm-theory}

\begin{figure}[ht!]
    \centering
    \includegraphics[width=1.0\linewidth]{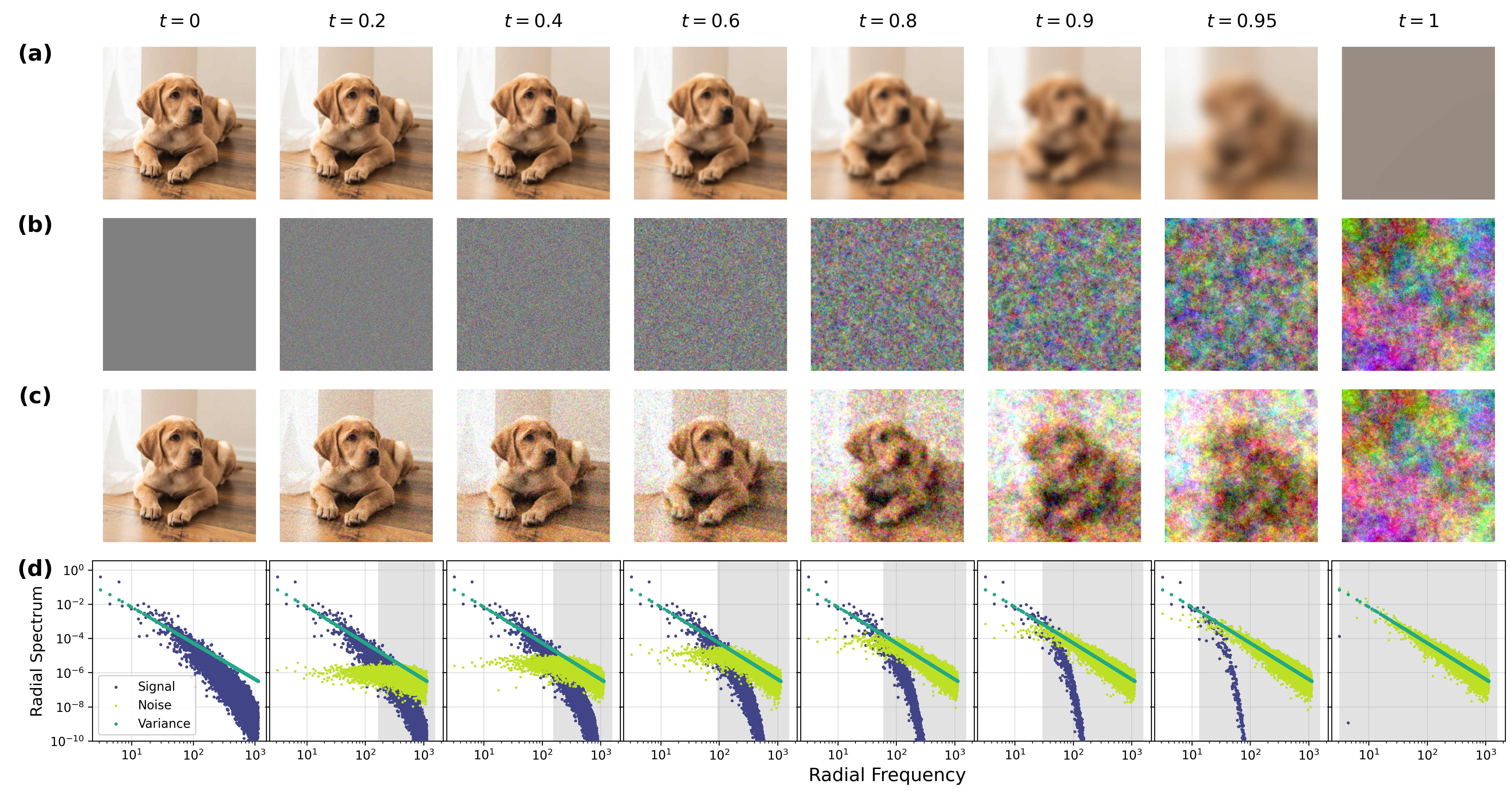}
    \caption{Forward trajectory of SKILD on a natural image.
    (a), (b), and (c) show the signal, noise, and full forward process,
    respectively. (d) shows the radial power spectrum evolution measured
    from the trajectory. High frequencies are suppressed first, while
    low-frequency content persists until late timesteps. Gray regions
    indicate modes below the SNR threshold, where resolution is effectively
    lost.}
    \label{fig:dog_forward}
\end{figure}

This appendix derives the DDPM posterior used in the main text and lists the
training targets supported by our implementation.
Figure~\ref{fig:dog_forward} shows a forward trajectory on a natural image,
illustrating how high-frequency content is removed before low-frequency
content under the scale-invariant schedule.

\subsection{Posterior derivation}

The forward marginal in main text Eq.~\eqref{eq:kddpm} can also be written as a
one-step Markov transition,
\begin{equation}\label{eq:app-onestep}
    \mathbf{X}_n
    =
    \sqrt{\ba_n} \odot \mathbf{X}_{n-1}
    +
    \sqrt{\bb_n} \odot \sqrt{\mathbf{S}_0}\,\epsilon_n,
    \qquad
    \epsilon_n \sim \mathcal{N}(0, \mathbf{I}),
\end{equation}
with $\bar{\ba}_n = e^{-\mathbf{k}^2 \lambda_n}$, $\ba_n = \bar{\ba}_n /
\bar{\ba}_{n-1}$, $\bb_n = 1 - \ba_n$, and $\bar{\ba}_0 = \mathbf{1}$. Note
that $\bar{\ba}_n = \prod_{i=1}^n \ba_i$.

When $\mathbf{X}_0$ is known, the reverse posterior $q(\mathbf{X}_{n-1} \mid
\mathbf{X}_n, \mathbf{X}_0)$ is exact. By Bayes' theorem,
\begin{equation}
    q(\mathbf{X}_{n-1} \mid \mathbf{X}_n, \mathbf{X}_0)
    \propto
    q(\mathbf{X}_n \mid \mathbf{X}_{n-1})\, q(\mathbf{X}_{n-1} \mid \mathbf{X}_0),
\end{equation}
where we used the Markov property and dropped the
$\mathbf{X}_{n-1}$-independent factor $q(\mathbf{X}_n \mid \mathbf{X}_0)$.
Eq.~\eqref{eq:app-onestep} gives
\begin{equation}
    q(\mathbf{X}_n \mid \mathbf{X}_{n-1})
    = \mathcal{N}\!\left(\mathbf{X}_n;\, \sqrt{\ba_n} \odot \mathbf{X}_{n-1},\,
                          \mathbf{S}_0 \bb_n\right)
    \;\propto\;
    \mathcal{N}\!\left(\mathbf{X}_{n-1};\, \mathbf{X}_n / \sqrt{\ba_n},\,
                          \mathbf{S}_0 \bb_n / \ba_n\right),
\end{equation}
where the second form completes the square in $\mathbf{X}_{n-1}$. The forward
marginal at $n-1$ is
\begin{equation}
    q(\mathbf{X}_{n-1} \mid \mathbf{X}_0)
    = \mathcal{N}\!\left(\mathbf{X}_{n-1};\, \sqrt{\bar{\ba}_{n-1}}
                          \odot \mathbf{X}_0,\, \mathbf{S}_0 (1 - \bar{\ba}_{n-1})\right).
\end{equation}
Multiplying these two Gaussians and using the standard product formula
$\tilde{\mathbf{v}} = (\mathbf{v}_1^{-1} + \mathbf{v}_2^{-1})^{-1}$,
$\tilde{\mathbf{m}} = \tilde{\mathbf{v}}(\mathbf{m}_1 / \mathbf{v}_1
+ \mathbf{m}_2 / \mathbf{v}_2)$ gives
\begin{subequations}\label{eq:app-posterior}
    \begin{align}
        q(\mathbf{X}_{n-1} \mid \mathbf{X}_n, \mathbf{X}_0)
        &= \mathcal{N}\!\left(\boldsymbol{\mu}_q,\,
                                \mathbf{S}_0 \tilde{\bb}_n\right), \\
        \tilde{\bb}_n
        &= \frac{\bb_n (1 - \bar{\ba}_{n-1})}{1 - \bar{\ba}_n}, \\
        \boldsymbol{\mu}_q
        &= \frac{\sqrt{\bar{\ba}_{n-1}}\, \bb_n}{1 - \bar{\ba}_n}\,
            \mathbf{X}_0
        + \frac{\sqrt{\ba_n}\,(1 - \bar{\ba}_{n-1})}{1 - \bar{\ba}_n}\,
            \mathbf{X}_n.
    \end{align}
\end{subequations}
Reverse sampling with the exact posterior reads
\begin{equation}\label{eq:app-ddpm-sample}
    \mathbf{X}_{n-1}
    = \boldsymbol{\mu}_q(n, \mathbf{X}_n, \mathbf{X}_0)
    + \sqrt{\mathbf{S}_0 \tilde{\bb}_n} \odot \epsilon_n.
\end{equation}

\subsection{Training targets}

At sampling time, $\mathbf{X}_0$ is unknown and is replaced by a network
estimate $\mathbf{X}_{0, \theta}$ inserted into Eq.~\eqref{eq:app-posterior}.
Our implementation supports four prediction targets: noise $\epsilon$, data
$\mathbf{X}_0$, an unwhitened frequency-space velocity $\mathbf{w}$, and a
whitened velocity $\mathbf{v}$. The main results use $\epsilon$-prediction;
the other targets follow the same posterior algebra and are reported as
diagnostic ablations.

\paragraph{$\epsilon$-prediction.}
Minimizing the MSE against the noise variable,
\begin{equation}
    \mathcal{L}
    = \mathbb{E}_{n, \mathbf{X}_0, \epsilon}\!
    \left[\lVert \epsilon - \epsilon_\theta(\mathbf{X}_n, n) \rVert_2^2\right],
\end{equation}
inverts main text Eq.~\eqref{eq:kddpm} to obtain
\begin{equation}
    \mathbf{X}_{0, \theta}
    = \frac{\mathbf{X}_n - \sqrt{1 - \bar{\ba}_n} \odot \sqrt{\mathbf{S}_0}\,
            \epsilon_{n,\theta}}{\sqrt{\bar{\ba}_n}},
\end{equation}
and substituting into Eq.~\eqref{eq:app-posterior} gives the $\epsilon$-form
of the posterior mean used at sampling,
\begin{equation}
    \boldsymbol{\mu}_\theta(n, \mathbf{X}_n)
    = \frac{1}{\sqrt{\ba_n}}
    \left(
        \mathbf{X}_n
        - \frac{\bb_n \sqrt{\mathbf{S}_0}}{\sqrt{1 - \bar{\ba}_n}}\,
          \epsilon_{n,\theta}
    \right).
\end{equation}
In finite precision we floor very small $\ba_n(\mathbf{k})$ values during
sampling, as described in the main text, to avoid unstable divisions in
high-frequency modes.

\paragraph{$\mathbf{X}_0$-prediction.}
Minimizing the MSE against the data,
\begin{equation}
    \mathcal{L}
    = \mathbb{E}_{n, \mathbf{X}_0, \epsilon}\!
    \left[\lVert \mathbf{X}_0 - \mathbf{X}_{0,\theta}(\mathbf{X}_n, n)
                 \rVert_2^2\right],
\end{equation}
substitutes $\mathbf{X}_{0,\theta}$ directly into the posterior mean,
\begin{equation}
    \boldsymbol{\mu}_\theta(n, \mathbf{X}_n)
    = \frac{\sqrt{\bar{\ba}_{n-1}}\, \bb_n}{1 - \bar{\ba}_n}\,
        \mathbf{X}_{0, \theta}
    + \frac{\sqrt{\ba_n}\,(1 - \bar{\ba}_{n-1})}{1 - \bar{\ba}_n}\,
        \mathbf{X}_n.
\end{equation}

\paragraph{$\mathbf{w}$-prediction.}
Let $a = \sqrt{\bar{\ba}_n}$ and $b = \sqrt{1 - \bar{\ba}_n}$. The forward
marginal and the velocity target are
\begin{subequations}
    \begin{align}
        \mathbf{X}_n
        &= a \odot \mathbf{X}_0
        + b \odot \sqrt{\mathbf{S}_0}\,\epsilon_n, \\
        \mathbf{w}_n
        &= a \odot \sqrt{\mathbf{S}_0}\,\epsilon_n
        - b \odot \mathbf{X}_0.
    \end{align}
\end{subequations}
Given a prediction $\mathbf{w}_{n,\theta}$, the data and colored-noise
estimates follow without division (using $a^2 + b^2 = 1$):
\begin{subequations}
    \begin{align}
        \mathbf{X}_{0, \theta}
        &= a \odot \mathbf{X}_n - b \odot \mathbf{w}_{n, \theta}, \\
        \mathbf{z}_{n, \theta}
        &= b \odot \mathbf{X}_n + a \odot \mathbf{w}_{n, \theta},
    \end{align}
\end{subequations}
where $\mathbf{z}_{n, \theta}$ estimates $\sqrt{\mathbf{S}_0} \odot \epsilon_n$.
The corresponding loss is
\begin{equation}
    \mathcal{L}_w
    = \mathbb{E}_{n, \mathbf{X}_0, \epsilon}\!
    \left[\lVert \mathbf{w}_n - \mathbf{w}_{n, \theta}(\mathbf{X}_n, n)
                 \rVert_2^2\right].
\end{equation}

\paragraph{$\mathbf{v}$-prediction.}
The whitened velocity divides $\mathbf{w}$ by $\sqrt{\mathbf{S}_0}$,
\begin{equation}
    \mathbf{v}_n
    = a \odot \epsilon_n
    - b \odot \mathbf{X}_0 / \sqrt{\mathbf{S}_0}.
\end{equation}
For $\mathbf{w}$ and $\mathbf{v}$, the loss can optionally be weighted by the
local frequency shell $\bar{\ba}_{n-1} - \bar{\ba}_n$. In our ablations, direct
prediction of the colored noise $\sqrt{1 - \bar{\ba}_n} \odot \sqrt{\mathbf{S}_0}\,
\epsilon_n$ matches the CIFAR-10 FID and IS of $\epsilon$-prediction but
converges more slowly, while $\mathbf{X}_0$- and $\mathbf{w}$-prediction
converge more slowly still and were not used for the reported metrics.

\section{SDE formulation}\label{app:sde-theory}

\subsection{Forward and reverse SDEs}\label{app:sde}

Following the score-matching formulation of \cite{song2020score}, we write a
forward stochastic differential equation (SDE) and its time-reverse counterpart
as
\begin{subequations}
    \begin{align}
        \text{forward: }\,
        \mathrm{d}\mathbf{X}
        &= \mathbf{f}(\mathbf{X}, t)\, \mathrm{d}t
        + \mathbf{g}(t) \odot \mathrm{d}\mathbf{w},
        \label{eq:forward-general} \\
        \text{reverse: }\,
        \mathrm{d}\mathbf{X}
        &= \left[
            \mathbf{f}(\mathbf{X}, t)
            - \mathbf{g}^2(t) \odot \nabla_{\mathbf{X}} \log p_t(\mathbf{X})
        \right] \mathrm{d}t
        + \mathbf{g}(t) \odot \mathrm{d}\bar{\mathbf{w}}.
        \label{eq:reverse-general}
    \end{align}
\end{subequations}
The whole process is computed in frequency space. We use the main text
notation: $\mathbf{X}$ for DCT coefficients, $\mathbf{S}_0$ for the empirical
spectrum, and $\epsilon \sim \mathcal{N}(0, \mathbf{I})$ for normalized
Gaussian noise. We suppress explicit $\mathbf{k}$ arguments where no ambiguity
arises.

The continuous-time SDE can be obtained either by differentiating
Eq.~\eqref{eq:field-eq} or, more compactly, by starting from a linear SDE
ansatz, integrating it exactly, and matching the resulting mean and variance
to the desired forward marginal. We follow the second route, which determines
$\mathbf{f}(\mathbf{X}, t)$ and $\mathbf{g}(t)$ mode by mode.

Since $\mathbf{X}$ is evolved under a Gaussian filter in
Eq.~\eqref{eq:field-eq}, the SDE ansatz should be linear in $\mathbf{X}$. For
each mode $\mathbf{k}$, write
\begin{equation}\label{eq:ansatz}
    \mathrm{d}x = -a(t)\, x\, \mathrm{d}t + b(t)\,\mathrm{d}w.
\end{equation}
Define the integrating factor $m(t) \equiv \exp\!\left(\int_0^t a(u)\,
\mathrm{d}u\right)$, so that $\mathrm{d}m = a(t) m(t)\,\mathrm{d}t$. The Itô
product rule gives
\begin{equation}
    \mathrm{d}(m x)
    = m\,\mathrm{d}x + x\,\mathrm{d}m + \mathrm{d}[m, x]
    = m\, b(t)\,\mathrm{d}w,
\end{equation}
since $m$ has finite variation and so $\mathrm{d}[m, x] = 0$. Integrating and
dividing by $m(t)$, with $m(0) = 1$,
\begin{equation}
    x_t
    = \underbrace{e^{-\int_0^t a(u)\,\mathrm{d}u}\, x_0}_{\text{signal}}
    + \underbrace{\int_0^t e^{-\int_s^t a(u)\,\mathrm{d}u}\, b(s)\,
                  \mathrm{d}w_s}_{\text{noise}}.
\end{equation}
Matching $\mathbb{E}[x_t]$ in this expression to the forward marginal
Eq.~\eqref{eq:field-eq} gives
\begin{equation}
    e^{-\int_0^t a(u)\,\mathrm{d}u}\,\mathbb{E}[x_0]
    \stackrel{!}{=} e^{-k^2 \lambda_t / 2}\,\mathbb{E}[x_0]
    \quad\Longrightarrow\quad
    a(t) = \tfrac{1}{2}\, k^2\, \dot\lambda(t).
\end{equation}
Matching $\operatorname{Var}[x_t]$ requires $\operatorname{Var}[x_t] = S_0$ at
every $t$ when $\operatorname{Var}[x_0] = S_0$, the dataset spectrum:
\begin{equation}
    e^{-2\int_0^t a\,\mathrm{d}u}\, S_0
    + \int_0^t e^{-2\int_s^t a\,\mathrm{d}u}\, b^2(s)\,\mathrm{d}s
    \stackrel{!}{=} S_0.
\end{equation}
Differentiating with respect to $t$ and substituting,
\begin{equation}
    -k^2 \dot\lambda_t\, S_0\,(1 - e^{-k^2 \lambda_t})
    + b^2(t)
    = S_0\, k^2\, \dot\lambda_t\, e^{-k^2 \lambda_t}
    \quad\Longrightarrow\quad
    b^2(t) = k^2\, \dot\lambda_t\, S_0.
\end{equation}
Restoring all $\mathbf{k}$ modes, we obtain
\begin{subequations}
    \begin{align}
        \mathbf{f}(\mathbf{X}, t)
        &= -\tfrac{1}{2}\, \mathbf{k}^2\, \dot\lambda(t)\, \mathbf{X}_t, \\
        \mathbf{g}(t)
        &= \sqrt{\mathbf{k}^2\, \dot\lambda(t)\, \mathbf{S}_0},
    \end{align}
\end{subequations}
and the forward and reverse SDEs read
\begin{subequations}
    \begin{align}
        \text{forward: }\,
        \mathrm{d}\mathbf{X}_t
        &= -\tfrac{1}{2}\, \mathbf{k}^2\, \dot\lambda(t)\, \mathbf{X}_t\,
            \mathrm{d}t
        + \sqrt{\mathbf{k}^2\, \dot\lambda(t)\, \mathbf{S}_0}
            \odot \mathrm{d}\mathbf{w},
        \label{eq:forward} \\
        \text{reverse: }\,
        \mathrm{d}\mathbf{X}_t
        &= \left[
            -\tfrac{1}{2}\, \mathbf{k}^2\, \dot\lambda(t)\, \mathbf{X}_t
            - \mathbf{k}^2\, \dot\lambda(t)\, \mathbf{S}_0
              \odot \nabla_{\mathbf{X}} \log p_t(\mathbf{X})
        \right] \mathrm{d}t
        + \sqrt{\mathbf{k}^2\, \dot\lambda(t)\, \mathbf{S}_0}
            \odot \mathrm{d}\bar{\mathbf{w}}.
        \label{eq:reverse}
    \end{align}
\end{subequations}

\subsection{Score function and training objectives}\label{app:sde-train}

To determine the training target for a score network, we compute the marginal
score. Let $\mathbf{A}_t = e^{-\mathbf{k}^2 \lambda(t) / 2}$ and $\mathbf{D}_t
= \sqrt{\mathbf{S}_0 (1 - e^{-\mathbf{k}^2 \lambda(t)})}$ denote the signal and
noise prefactors. The forward conditional is $\mathcal{N}(\mathbf{X};\,
\mathbf{A}_t \odot \mathbf{X}_0,\, \mathbf{D}_t^2)$, and the marginal is
\begin{equation}
    p_t(\mathbf{X})
    = \int p_0(\mathbf{X}_0)\,
      \mathcal{N}(\mathbf{X};\, \mathbf{A}_t \odot \mathbf{X}_0,\,
                  \mathbf{D}_t^2)\,
      \mathrm{d}\mathbf{X}_0.
\end{equation}
Differentiating under the integral and dividing by $p_t(\mathbf{X})$ gives
the score
\begin{equation}\label{eq:score}
    \nabla_{\mathbf{X}} \log p_t(\mathbf{X})
    = \mathbb{E}\!\left[\mathbf{D}_t^{-2} \odot
                        (\mathbf{A}_t \odot \mathbf{X}_0 - \mathbf{X})
                        \,\big|\, \mathbf{X}\right]
    = \mathbf{D}_t^{-2} \odot
      \left(\mathbf{A}_t \odot \mathbb{E}[\mathbf{X}_0 \mid \mathbf{X}]
            - \mathbf{X}\right),
\end{equation}
which gives the Tweedie identity
\begin{equation}
    \mathbb{E}[\mathbf{X}_0 \mid \mathbf{X}]
    = \mathbf{A}_t^{-1}
      \odot \left(\mathbf{X}
                  + \mathbf{D}_t^2 \odot \nabla_{\mathbf{X}} \log p_t(\mathbf{X})
            \right).
\end{equation}

The standard score-matching objective is
\begin{equation}\label{eq:loss}
    \mathcal{L}
    = \mathbb{E}_{t, \mathbf{X}_0, \epsilon}\!\left[
        \boldsymbol{\omega}(t) \odot
        \lVert \mathbf{s}_\theta(\mathbf{X}, t)
                - \nabla_{\mathbf{X}} \log p_t(\mathbf{X})
        \rVert_2^2
    \right]
    = \mathbb{E}_{t, \mathbf{X}_0, \epsilon}\!\left[
        \boldsymbol{\omega}(t) \odot
        \lVert \mathbf{s}_\theta(\mathbf{X}, t)
                + \mathbf{D}_t^{-1} \odot \epsilon_t
        \rVert_2^2
    \right],
\end{equation}
where $\mathbf{s}_\theta$ is the score network and $\boldsymbol{\omega}(t)$ is
a per-mode weighting function.\footnote{Strictly, the loss should reference
$\nabla_{\mathbf{X}} \log p_t(\mathbf{X} \mid \mathbf{X}_0)$. The two
objectives differ by an $\mathbf{s}_\theta$-independent constant via the
Fisher identity $\mathbb{E}[\nabla_{\mathbf{X}} \log p_t(\mathbf{X} \mid
\mathbf{X}_0) \mid \mathbf{X}] = \nabla_{\mathbf{X}} \log p_t(\mathbf{X})$,
so they share the same minimizer. This is the basis of denoising score
matching.} The regression target $-\mathbf{D}_t^{-1} \odot \epsilon_t$ is the
anisotropic analogue of the variance-preserving (VP) diffusion target.
Without a low-frequency cutoff, the DC mode $\mathbf{k} = 0$ would be
unchanged because neither drift nor noise act on it; in practice we replace
$\lVert \mathbf{k} \rVert$ by $\max(\lVert \mathbf{k} \rVert, k_c)$ for small
modes, so the same equations apply with a nonzero low-frequency cutoff.

The choice of $\boldsymbol{\omega}(t)$ determines the prediction target.
Three natural choices recover the targets discussed in
Appendix~\ref{app:ddpm-theory}.

\paragraph{$\epsilon$-net weight.}
Setting $\boldsymbol{\omega}(t) \propto \mathbf{D}_t^2 = \mathbf{S}_0
(1 - e^{-\mathbf{k}^2 \lambda(t)})$ and writing $\mathbf{s}_\theta
= -\mathbf{D}_t^{-1} \odot \epsilon_\theta$, the loss reduces to
\begin{equation}
    \mathcal{L}
    = \mathbb{E}_{t, \mathbf{X}_0, \epsilon}\!\left[
        \lVert \epsilon_\theta(\mathbf{X}, t) - \epsilon_t \rVert_2^2
    \right].
\end{equation}

\paragraph{$\mathbf{X}_0$-net weight.}
Setting $\boldsymbol{\omega}(t) \propto \mathbf{A}_t^{-2} \odot \mathbf{D}_t^4$
and writing $\mathbf{s}_\theta = \mathbf{D}_t^{-2} \odot
(\mathbf{A}_t \odot \mathbf{X}_{0, \theta} - \mathbf{X})$ via Tweedie, the loss
reduces to
\begin{equation}
    \mathcal{L}
    = \mathbb{E}_{t, \mathbf{X}_0, \epsilon}\!\left[
        \lVert \mathbf{X}_{0, \theta}(\mathbf{X}, t) - \mathbf{X}_0
        \rVert_2^2
    \right].
\end{equation}

\paragraph{KL-minimizing weight.}
Setting $\boldsymbol{\omega}(t) \propto \mathbf{g}^2(t)$ minimizes the
Kullback-Leibler (KL) divergence between the true and learned reverse
dynamics over the whole path: a score error $\mathbf{e}(\mathbf{k})$ produces
drift error $\mathbf{g}(t) \odot \mathbf{e}(\mathbf{k})$, and for SDEs
sharing the same diffusion the path-KL scales as $\tfrac{1}{2} \int
\mathbf{g}^2(t) \odot \mathbb{E}[\mathbf{e}(\mathbf{k})^2]\,\mathrm{d}t$.
Modes and times where the reverse drift amplifies score errors then receive
proportionally more weight during training.

\subsection{Samplers}\label{app:sde-sample}

The standard score-matching samplers \cite{song2020score} include the
Euler-Maruyama (EM) sampler, the Ordinary Differential Equation (ODE) sampler
from probability flows, and the predictor-corrector (PC) sampler. Our
experiments use ancestral DDPM sampling
(Eq.~\eqref{eq:app-ddpm-sample}); the SDE samplers below are reference forms.

\paragraph{EM sampler.}
On a decreasing time grid $1 = t_N > t_{N-1} > \cdots > t_0 = 0$ with
$\Delta t_n = t_{n-1} - t_n < 0$, discretizing Eq.~\eqref{eq:reverse} gives
\begin{equation}
    \mathbf{X}_{n-1}
    = \mathbf{X}_n
    + \left[
        -\tfrac{1}{2}\, \mathbf{k}^2\, \dot\lambda_n\, \mathbf{X}_n
        - \mathbf{k}^2\, \dot\lambda_n\, \mathbf{S}_0
          \odot \mathbf{s}_\theta(\mathbf{X}_n, t_n)
    \right] \Delta t_n
    + \sqrt{\mathbf{k}^2\, \dot\lambda_n\, \mathbf{S}_0}
      \odot \sqrt{-\Delta t_n}\,\epsilon_n.
\end{equation}

\paragraph{ODE sampler.}
For any SDE of the form Eq.~\eqref{eq:forward-general}, the probability flow
ODE
\begin{equation}
    \mathrm{d}\mathbf{X}
    = \left[
        \mathbf{f}(\mathbf{X}, t)
        - \mathbf{g}^2(t) \odot \nabla_{\mathbf{X}} \log p_t(\mathbf{X})
    \right] \mathrm{d}t
\end{equation}
matches the SDE marginals $p_t(\mathbf{X})$. Solving in reverse time yields
the same target distribution as the reverse SDE, with the Brownian term
dropped:
\begin{equation}
    \mathbf{X}_{n-1}
    = \mathbf{X}_n
    + \left[
        -\tfrac{1}{2}\, \mathbf{k}^2\, \dot\lambda_n\, \mathbf{X}_n
        - \mathbf{k}^2\, \dot\lambda_n\, \mathbf{S}_0
          \odot \mathbf{s}_\theta(\mathbf{X}_n, t_n)
    \right] \Delta t_n.
\end{equation}

\paragraph{PC sampler.}
The PC sampler refines an EM prediction $\mathbf{X}_{n-1}^p$ via a few
iterations of a score-based MCMC corrector at the same time level. A
preconditioned Langevin update that leaves $p_t(\mathbf{X})$ invariant is
\begin{equation}
    \mathrm{d}\mathbf{X}_\tau
    = \mathbf{B}(t) \odot \nabla_{\mathbf{X}} \log p_t(\mathbf{X})\,
      \mathrm{d}\tau
    + \sqrt{2\, \mathbf{B}(t)} \odot \mathrm{d}\mathbf{w}_\tau,
\end{equation}
which classically uses $\mathbf{B}(t) = \mathbf{I}$ but admits an anisotropic,
time-dependent choice given the mode-dependent diffusion. Discretizing with
step $\eta_n$ gives the corrector update
\begin{equation}
    \mathbf{X}_{i+1}
    = \mathbf{X}_i
    + \mathbf{B}(t_n) \odot \mathbf{s}_\theta(\mathbf{X}_n, t_n)\, \eta_n
    + \sqrt{2\, \mathbf{B}(t_n)\, \eta_n} \odot \epsilon_n,
\end{equation}
and one full PC step combines the EM predictor with $K$ Langevin corrections
initialized at $\mathbf{X}_{i=0} = \mathbf{X}_{n-1}^p$. The corrector
reduces the discretization error of the predictor.

%% file: sections/10_2_app_exp.tex
\section{Scale invariance and power laws in nature and
physics}\label{app:spectrum}

\subsection{Dictionary between physics and natural images}

We provide a brief analogy between field theory and natural images that
motivates treating them in the same framework.

A Monte Carlo sample of a scalar field theory can be encoded directly as an
image: each pixel at lattice site $\mathbf{r}$ stores the value of the field
variable at that site. The full distribution over field configurations is then
represented by the dataset of all such images. Conversely, a natural image can
be viewed as a single sample from a three-channel scalar field theory, with each
pixel storing a field value, the resolution playing the role of the inverse
lattice spacing, and the image size playing the role of the physical system
size. The infinite-resolution limit corresponds to the continuum limit of the
field, and the infinite-size limit corresponds to the thermodynamic limit. The
basic dictionary is
\begin{itemize}[itemsep=0.2em, topsep=0.2em]
    \item pixel value $\leftrightarrow$ field variable;
    \item image resolution $\leftrightarrow$ inverse lattice spacing (continuum
    limit at infinite resolution);
    \item image size $\leftrightarrow$ system size (thermodynamic limit at
    infinite size).
\end{itemize}
A critical field theory is scale-invariant under renormalization-group
coarse-graining and rescaling. The corresponding question for images is whether
a sufficiently large and high-resolution natural-image dataset is approximately
scale-invariant under the analogous operation of low-pass filtering followed by
zooming in. The variance spectra in Section~\ref{sec:prelims} indicate that, in
expectation across a dataset, the answer is approximately yes.

\subsection{Power-law benchmark in frequency space}

We specify the DCT, IDCT, and frequency normalization used throughout the paper,
then summarize the per-dataset variance fits.

\paragraph{DCT and IDCT conventions.}
The forward transform is the type-II DCT and the inverse is the type-III IDCT.
For an image $x \in \mathbb{R}^{H \times W}$,
\begin{equation}
    X_{u,v}
    =
    \frac{4}{HW}
    \sum_{i=0}^{H-1}\sum_{j=0}^{W-1}
    x_{i,j}
    \cos\!\left[\tfrac{\pi}{H}\!\left(i+\tfrac12\right)u\right]
    \cos\!\left[\tfrac{\pi}{W}\!\left(j+\tfrac12\right)v\right],
\end{equation}
with inverse
\begin{equation}
    x_{i,j}
    =
    \sum_{u,v}
    \gamma_u \gamma_v X_{u,v}
    \cos\!\left[\tfrac{\pi}{H}\!\left(i+\tfrac12\right)u\right]
    \cos\!\left[\tfrac{\pi}{W}\!\left(j+\tfrac12\right)v\right],
\end{equation}
where $\gamma_0 = 1/2$ and $\gamma_{k>0} = 1$.

\begin{figure}[t]
    \begin{minipage}{0.55\textwidth}
        \centering
        \begin{subfigure}{0.48\linewidth}
            \centering
            \includegraphics[width=\linewidth]{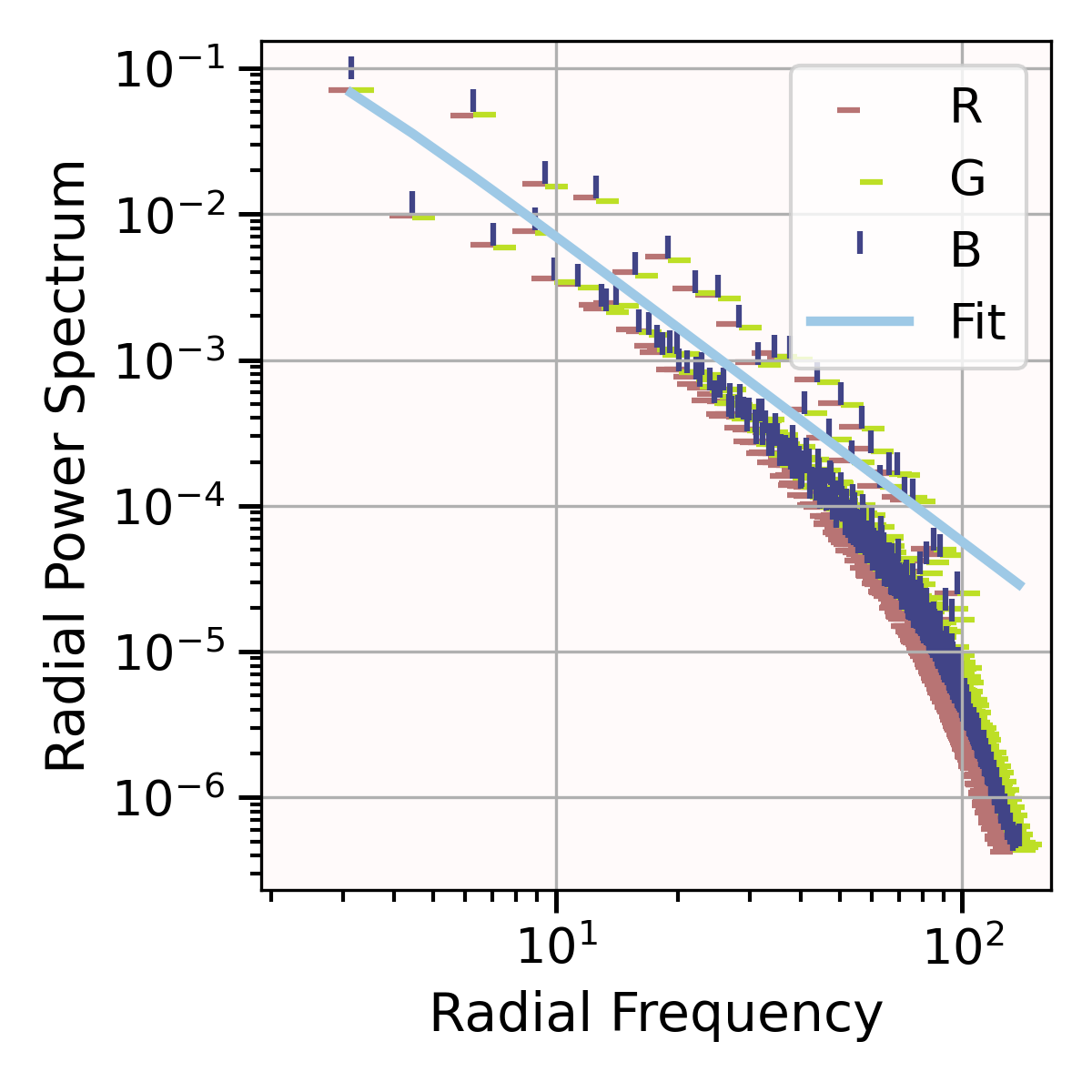}
            \caption{CIFAR-10 variance}
            \label{fig:cifar_variance}
        \end{subfigure}
        \hfill
        \begin{subfigure}{0.48\linewidth}
            \centering
            \includegraphics[width=\linewidth]{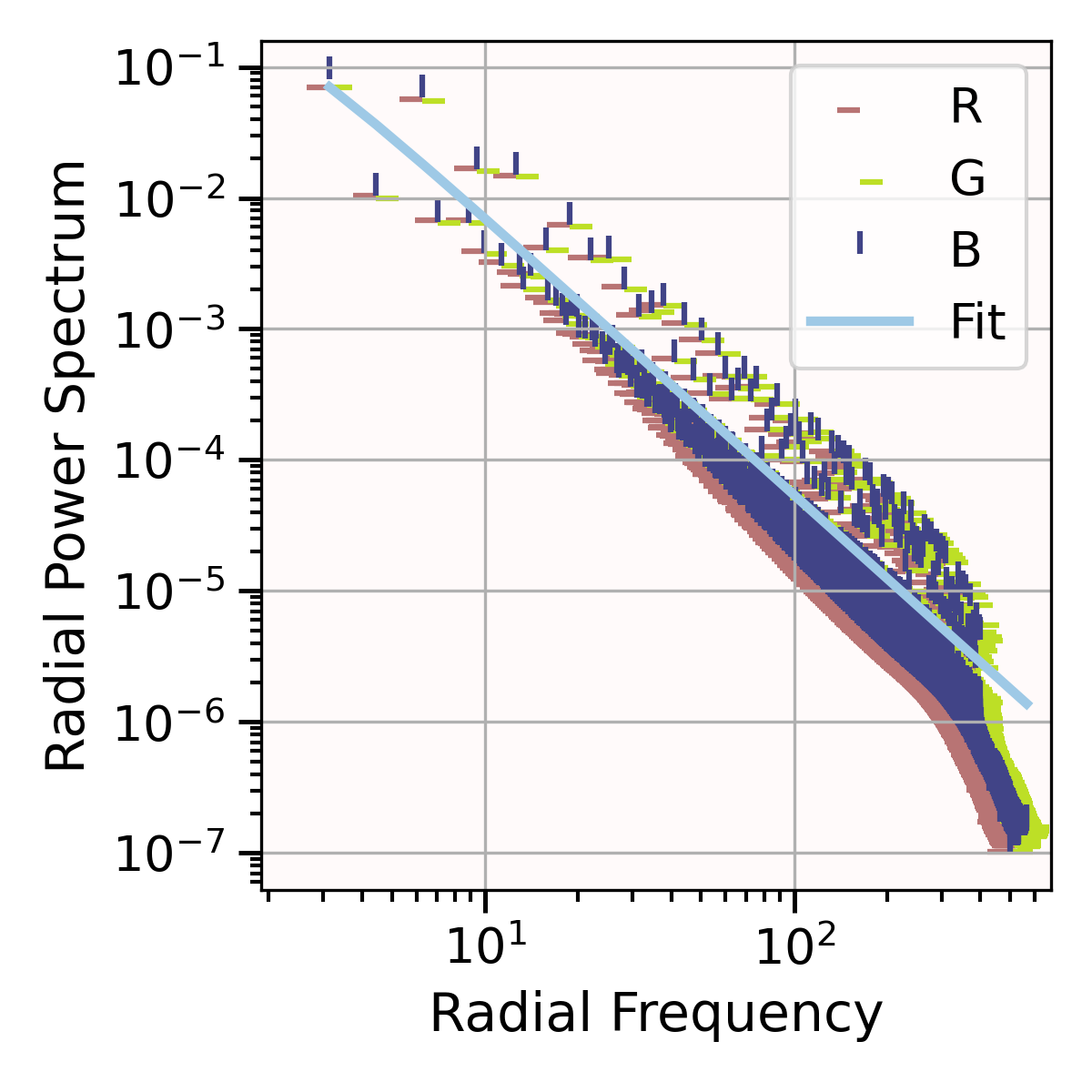}
            \caption{ImageNet-128 variance}
            \label{fig:imagenet128_variance}
        \end{subfigure}
    \end{minipage}%
    \hfill
    \begin{minipage}{0.4\textwidth}
        \captionsetup{justification=raggedright,singlelinecheck=false}
        \caption{Variance spectra for (a) CIFAR-10 and (b) ImageNet-128,
        computed independently for each color channel (RGB). Both power-law fits
        recover the $k^{-2}$ frequency decay reported in previous work. The
        deviations at the highest frequencies in (a) reflect finite-resolution
        limitations and shrink in (b) as resolution increases.}
        \label{fig:variance_supp}
    \end{minipage}
\end{figure}

\paragraph{Frequency normalization.}
We define the frequency vector by $\mathbf{k} = (\pi u, \pi v)$, so that mode
indices are spaced uniformly by $\pi$ and the magnitude $\lVert \mathbf{k}
\rVert$ ranges in $[0, \sqrt{2}\,\pi(H-1)]$ for an $H \times H$ image. Two
properties motivate this convention. First, the $4/(HW)$ amplitude factor in the
DCT keeps the maximum of the radial power spectrum at the same order of
magnitude across resolutions, as seen in Figure~\ref{fig:variance_supp} and main
text Figure~\ref{fig:variance_summary}. Second, the spacing-fixed convention
means that an image of higher resolution extends the spectrum to higher
$\mathbf{k}$ rather than rescaling existing modes; in the infinite-resolution
limit, the discrete grid fills out the continuum.

\paragraph{Variance fits.}
Figure~\ref{fig:variance_supp} complements the ImageNet-256 fit shown in main
text Figure~\ref{fig:variance_summary}(b) with the corresponding fits for
CIFAR-10 and ImageNet-128. All three recover an approximate $k^{-2}$ decay
across the shared frequency range; deviations at the highest frequencies reflect
the finite-resolution cutoff and shrink as resolution grows.
Table~\ref{tab:powerlaw_params} summarizes the fitted parameters.

\begin{table}[t]
    \centering
    \small
    \caption{Power-law fits of variance power spectra for the natural-image
    datasets.}
    \vspace{0.1 in}
    \label{tab:powerlaw_params}
    \begin{tabular}{lccc}
    \toprule
    Dataset & $C$ & $\mathbf{k}_0^2$ & $a$ \\
    \midrule
    CIFAR-10 & $0.9100 \pm 0.0182$ & $1.9406 \pm 0.0481$ & $1.0513 \pm 0.0021$
    \\
    ImageNet-128 & $0.9281 \pm 0.0016$ & $1.5708 \pm 0.0030$ & $1.0590 \pm
    0.0002$ \\
    ImageNet-256 & $0.9322 \pm 0.0004$ & $1.5708 \pm 0.0008$ & $1.0598 \pm
    0.0000$ \\
    \bottomrule
    \end{tabular}
\end{table}

\section{Supplementary materials for CIFAR-10 experiments}\label{app:cifar}

\subsection{Architecture and training}\label{app:cifar-train}

CIFAR-10 generation uses a score U-Net backbone from the NCSN++ family
\cite{song2020score} with discrete DDPM positional embeddings and $8$ residual
blocks. We scan the base channel count from $128$ to $256$ and report the best
configuration. Models are trained with AdamW for $400$K steps with batch size
$128$, learning rate $2 \times 10^{-4}$, weight decay $0$, and EMA rate $0.999$.
Checkpoints are saved every $20$K steps. Each CIFAR-10 model is trained on a
single H100 or GH200 GPU.

{\subsection{Evaluation protocol}\label{app:cifar-eval}

\paragraph{Checkpoint selection.}
We evaluate every saved checkpoint on $50$K generated samples. Following the
standard convention from \cite{song2020score}, we report the checkpoint with the
lowest FID during training along with its corresponding IS, rather than
averaging across checkpoints. For the run in Table~\ref{tab:cifar10}, FID
reaches its minimum at $160$K steps. IS continues to climb at later checkpoints
and hits $10.03$ at $400$K steps, while FID rises to $3.87$.
Appendix~\ref{app:mode-collapse} shows the full training trajectory.

\paragraph{Sampling seeds.}
Table~\ref{tab:seeds} evaluates the $160$K-step checkpoint from
Table~\ref{tab:cifar10} across three random seeds, using $50$K samples per seed.
Seed $42$ gives an FID of $2.65$ and an IS of $9.63$, matching the values noted
in Appendix~\ref{app:mode-collapse}. Table~\ref{tab:cifar10} reports seed $44$,
which achieves the lowest FID. Averaged across the three seeds, the model
reaches an FID of $2.632 \pm 0.022$ and an IS of $9.656 \pm 0.033$.

\begin{table}[ht]
    \centering
    \small
    \caption{FID and IS for the CIFAR-10 checkpoint in Table~\ref{tab:cifar10}
    across three random seeds, using $50$K samples per seed. Per-seed IS
    uncertainty shows the standard deviation across splits, and the bottom row
    reports the mean and standard deviation across the three seeds.}
    \vspace{0.1in}
    \label{tab:seeds}
    \begin{tabular}{lcc}
    \toprule
    Seed & FID $\downarrow$ & IS $\uparrow$ \\
    \midrule
    42 & 2.652 & $9.628 \pm 0.117$ \\
    43 & 2.637 & $9.692 \pm 0.139$ \\
    44 & 2.608 & $9.647 \pm 0.093$ \\
    \midrule
    Mean & $2.632 \pm 0.022$ & $9.656 \pm 0.033$ \\
    \bottomrule
    \end{tabular}
\end{table}
\par}

\subsection{Noise schedules}

We describe the two schedule families tested on CIFAR-10 and motivate their
forms. The schedule $\lambda(t)$ controls how the damping cutoff in $\mathbf{k}$
moves with time: a frequency mode is effectively suppressed when $\mathbf{k}^2
\lambda(t) \approx \theta$ for some threshold $\theta$, so the damping front is
$\mathbf{k}_d(t) = \sqrt{\theta / \lambda(t)}$.

\paragraph{Log-linear schedule.}
In the continuum limit ($\Delta \mathbf{k} \to 0$), a damping that is uniform on
a logarithmic frequency scale respects scale invariance: equal time intervals
remove equal logarithmic ranges of modes. The corresponding schedule is
\begin{equation}
    \lambda(t) = 10^{\lambda_i + (\lambda_f - \lambda_i) t},
\end{equation}
which gives $\mathbf{k}_d(t) \propto 10^{-(\lambda_f - \lambda_i) t / 2}$,
log-linear in $t$.

\paragraph{Linear schedule.}
Real datasets are at finite resolution and finite size, neither at the continuum
limit nor the thermodynamic limit, so it is also reasonable to damp $\mathbf{k}$
linearly with time:
\begin{equation}
    \lambda(t) = \frac{\theta}{(\lambda_f (1-t) + \lambda_i)^2},
\end{equation}
under which $\mathbf{k}_d(t)$ moves linearly with $t$.

\paragraph{Boundary fix.}
Both schedules above have $\lambda(0) > 0$ and $\dot\lambda(0) > 0$, so
high-frequency modes are damped abruptly at $t = 0$. This discontinuity removes
fine details before the model can learn them. We mitigate it by multiplying both
schedules by $t$, ensuring $\lambda(0) = 0$ and a smooth onset. The schedules
used in all experiments are
\begin{align}
    \text{log-linear:}\quad
    &\lambda(t) = t \cdot 10^{\lambda_i + (\lambda_f - \lambda_i) t}, \\
    \text{linear:}\quad
    &\lambda(t) = \frac{\theta\, t}{(\lambda_f (1 - t) + \lambda_i)^2}.
\end{align}

\begin{figure}[t]
    \centering
    \begin{minipage}{0.6\textwidth}
        \centering
        \includegraphics[width=\linewidth]{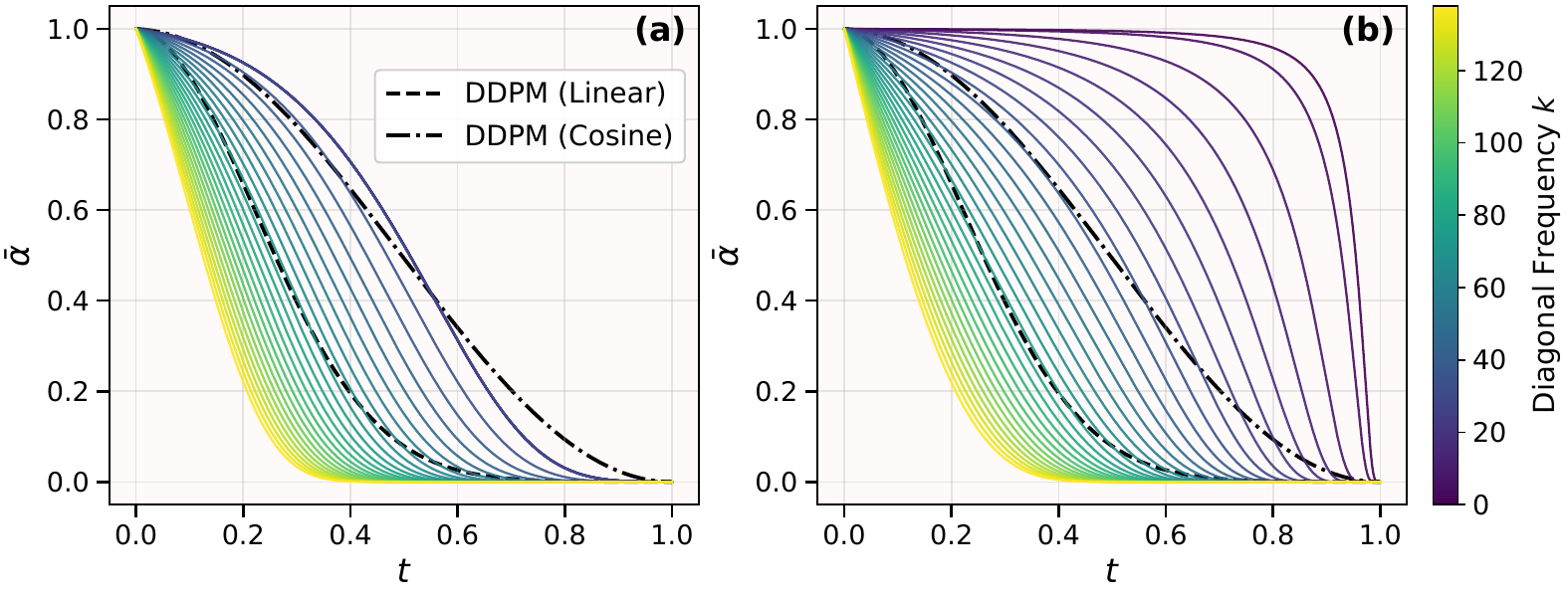}
    \end{minipage}%
    \hfill
    \begin{minipage}{0.38\textwidth}
        \captionsetup{justification=raggedright,singlelinecheck=false}
        \caption{Best (a) log-linear and (b) linear schedules for CIFAR-10
        generation. Different diagonal frequency modes are attenuated
        independently over time, with higher-frequency modes decaying before
        lower-frequency ones. The linear and cosine DDPM schedules are shown for
        reference; they are scalar in frequency space and attenuate all modes
        uniformly.}
        \label{fig:k-sched}
    \end{minipage}
\end{figure}

\paragraph{Best schedules on CIFAR-10.}
Figure~\ref{fig:k-sched} shows the best-performing schedule from each family.
The exact parameters of the best linear schedule, used to report the FID in the
main text, are $\theta = 5.0$, $\lambda_i = 137.7294$, $\lambda_f = 1.57$, $k_c
= 3$, and $N = 1000$. The best log-linear schedule uses $\lambda_i = -3.75$,
$\lambda_f = -2.0$, $k_c = 31.2$, and $N = 1000$.

\subsection{Ablation studies}\label{app:ablation}

\subsubsection{Mode collapse}\label{app:mode-collapse}

In every CIFAR-10 run without the fixes in Table~\ref{tab:lowfreq}, we
observe a clear mode collapse where FID reaches an early minimum and then rises
while IS continues to improve. For the linear schedule of
Figure~\ref{fig:k-sched}(b), FID converges to $2.65$ at step $160$K with IS
$9.63$, then degrades to $2.82$ at $200$K (IS $9.77$), $2.98$ at $250$K (IS
$9.79$), $3.26$ at $300$K (IS $9.98$), and $3.87$ at $400$K (IS $10.03$). Main
text Figure~\ref{fig:sched_ablation}(b) plots these trajectories.

We attribute this to instability in the low-frequency portion of the diffusion.
Late in the forward process, small changes in the surviving low-frequency modes
produce large overall color shifts in the noised image, visible in
Figures~\ref{fig:concept} and~\ref{fig:dog_forward}. This makes it harder for
the model to fix the global color tone and class-level structure of an image,
and the consequence is most pronounced on object-centric datasets such as
CIFAR-10. The pattern is consistent with image fidelity continuing to improve
(IS rising) while the marginal class distribution drifts away from the dataset
(FID degrading). The effect is strongest at large $k_c$, where many
low-frequency modes are bundled together at late times.

The low-frequency limit is not strictly scale-invariant, so we expect this
regime to need a different treatment. Plausible remedies include chaining the
SKILD reverse process with a small VAE or with a standard pixel-space diffusion
model that handles the very-low-frequency modes. The effect should also weaken
on higher-resolution and less object-centric natural-image datasets.
Table~\ref{tab:lowfreq} in the main text tests both this VAE approach and a
class-conditional model with classifier-free guidance. Both methods keep
training stable through $400$K steps.

\subsubsection{Schedule robustness}

Diffusion models are sensitive to noise schedules \cite{chen2023importance,
nichol2021improved}. Table~\ref{tab:schedule-robustness} reports a sweep over
$\lambda_i$, $\lambda_f$, $k_c$, and $\theta$ in both the log-linear and linear
families. Most schedules reach FID near or below $5$ within $400$K training
steps, and all reach IS at or above $9$. Convergence speed depends on the
schedule, but the final converged scores are not strongly schedule-dependent,
indicating that SKILD does not require a finely tuned schedule for stable
training.

\begin{table}[t]
  \centering
  \small
  \setlength{\tabcolsep}{4pt}
  \renewcommand{\arraystretch}{1.05}
  \caption{\textbf{Schedule robustness on CIFAR-10.} Each row reports one
  frequency-space diffusion schedule. Metrics are shown for $200$K and $400$K
  training steps. Across the completed $400$K evaluations, most schedules reach
  FID below or near $5$, and all reach Inception Score near or above $9$,
  indicating that SKILD is not tuned to a single fragile schedule. Convergence
  speed varies by schedule. A dash denotes an unused parameter or an unavailable
  run. The network used has half the channel count compared to the one in Table \ref{tab:cifar10} and Figure \ref{fig:sched_ablation}(b).}
  \vspace{0.1in}
  \label{tab:schedule-robustness}
  \resizebox{0.5\linewidth}{!}{%
  \begin{tabular}{@{}ccccrrrr@{}}
    \toprule
    \multicolumn{4}{c}{Schedule parameters}
    & \multicolumn{2}{c}{FID $\downarrow$} & \multicolumn{2}{c}{IS $\uparrow$}
    \\
    \cmidrule(lr){1-4} \cmidrule(lr){5-6} \cmidrule(l){7-8} $k_c$ & $\lambda_i$
    & $\lambda_f$ & $\theta$ & 200K & 400K & 200K & 400K \\
    \midrule
    \multicolumn{8}{@{}l}{\textit{Log-linear schedule}} \\
    \midrule
    $3.0$  & $-4.25$ & $0.0$  & -- & 6.77 & 5.53 & 8.72 & 8.98 \\
    $3.0$  & $-4.25$ & $0.4$  & -- & 6.80 & 5.63 & 8.63 & 8.84 \\
    $3.0$  & $-4.25$ & $0.8$  & -- & 7.24 & 5.66 & 8.56 & 8.85 \\
    $3.0$  & $-3.75$ & $0.0$  & -- & 5.65 & 4.47 & 8.77 & 9.07 \\
    $3.0$  & $-3.75$ & $0.4$  & -- & 6.10 & 4.83 & 8.74 & 9.04 \\
    $3.0$  & $-3.75$ & $0.8$  & -- & 6.56 & 5.17 & 8.73 & 8.99 \\
    $3.0$  & $-3.25$ & $0.0$  & -- & 5.72 & 4.84 & 8.78 & 9.07 \\
    $3.0$  & $-3.25$ & $0.4$  & -- & 6.02 & 4.77 & 8.77 & 9.06 \\
    $3.0$  & $-3.25$ & $0.8$  & -- & 6.30 & 4.77 & 8.66 & 8.98 \\
    $13.4$ & $-4.25$ & $-1.4$ & -- & 5.71 & 4.62 & 8.81 & 9.15 \\
    $13.4$ & $-4.25$ & $-1.0$ & -- & 5.97 & 4.88 & 8.80 & 9.06 \\
    $13.4$ & $-4.25$ & $-0.6$ & -- & 6.40 & 5.35 & 8.75 & 8.95 \\
    $13.4$ & $-3.75$ & $-1.4$ & -- & 4.90 & 4.40 & 8.95 & 9.22 \\
    $13.4$ & $-3.75$ & $-1.0$ & -- & 5.08 & 4.55 & 8.83 & 9.10 \\
    $13.4$ & $-3.75$ & $-0.6$ & -- & 5.57 & 4.62 & 8.76 & 8.99 \\
    $13.4$ & $-3.25$ & $-1.4$ & -- & 4.62 & 5.29 & 8.97 & 9.16 \\
    $13.4$ & $-3.25$ & $-1.0$ & -- & 5.21 & 4.84 & 8.87 & 9.13 \\
    $13.4$ & $-3.25$ & $-0.6$ & -- & 5.45 & 4.71 & 8.73 & 9.05 \\
    $31.2$ & $-4.25$ & $-2.2$ & -- & 5.40 & 4.63 & 8.86 & 9.14 \\
    $31.2$ & $-4.25$ & $-2.0$ & -- & 5.76 & 4.74 & 8.82 & 9.05 \\
    $31.2$ & $-4.25$ & $-1.6$ & -- & 6.00 & 4.93 & 8.78 & 8.97 \\
    $31.2$ & $-3.75$ & $-2.2$ & -- & 4.53 & 4.74 & 9.15 & 9.30 \\
    $31.2$ & $-3.75$ & $-1.6$ & -- & 5.39 & 4.64 & 8.91 & 9.05 \\
    $31.2$ & $-3.25$ & $-2.2$ & -- & 4.58 & 5.74 & 9.07 & 9.17 \\
    $31.2$ & $-3.25$ & $-2.0$ & -- & 4.76 & 5.41 & 8.97 & 9.15 \\
    $31.2$ & $-3.25$ & $-1.6$ & -- & 5.30 & 4.65 & 8.81 & 9.08 \\
    \midrule
    \multicolumn{8}{@{}l}{\textit{Linear schedule}} \\
    \midrule
    $3.0$  & $137.7294$ & $1.57$  & $3.0$ & 5.42 & 4.41 & 9.06 & 9.39 \\
    $3.0$  & $137.7294$ & $1.57$  & $5.0$ & 4.89 & 4.27 & 9.12 & 9.29 \\
    $3.0$  & $137.7294$ & $1.57$  & $7.0$ & 6.75 & 5.33 & 9.39 & 9.55 \\
    $13.4$ & $137.7294$ & $12.0$  & $3.0$ & 5.61 & 5.14 & 9.23 & 9.29 \\
    $13.4$ & $137.7294$ & $12.0$  & $5.0$ & 4.61 & 4.99 & 9.05 & 9.52 \\
    $13.4$ & $137.7294$ & $12.0$  & $7.0$ & 4.45 & --   & 9.07 & --   \\
    $31.2$ & $137.7294$ & $29.5$  & $3.0$ & 7.63 & 7.44 & 9.18 & 9.41 \\
    $31.2$ & $137.7294$ & $29.5$  & $5.0$ & 4.43 & 5.01 & 9.12 & 9.51 \\
    $31.2$ & $137.7294$ & $29.5$  & $7.0$ & 4.76 & 5.96 & 9.02 & 9.25 \\
    \bottomrule
  \end{tabular}%
  }
\end{table}

\subsubsection{Convergence against pixel-space schedules}

\begin{figure}[t]
\centering
\begin{minipage}{0.6\textwidth}
    \centering
    \includegraphics[width=\linewidth]{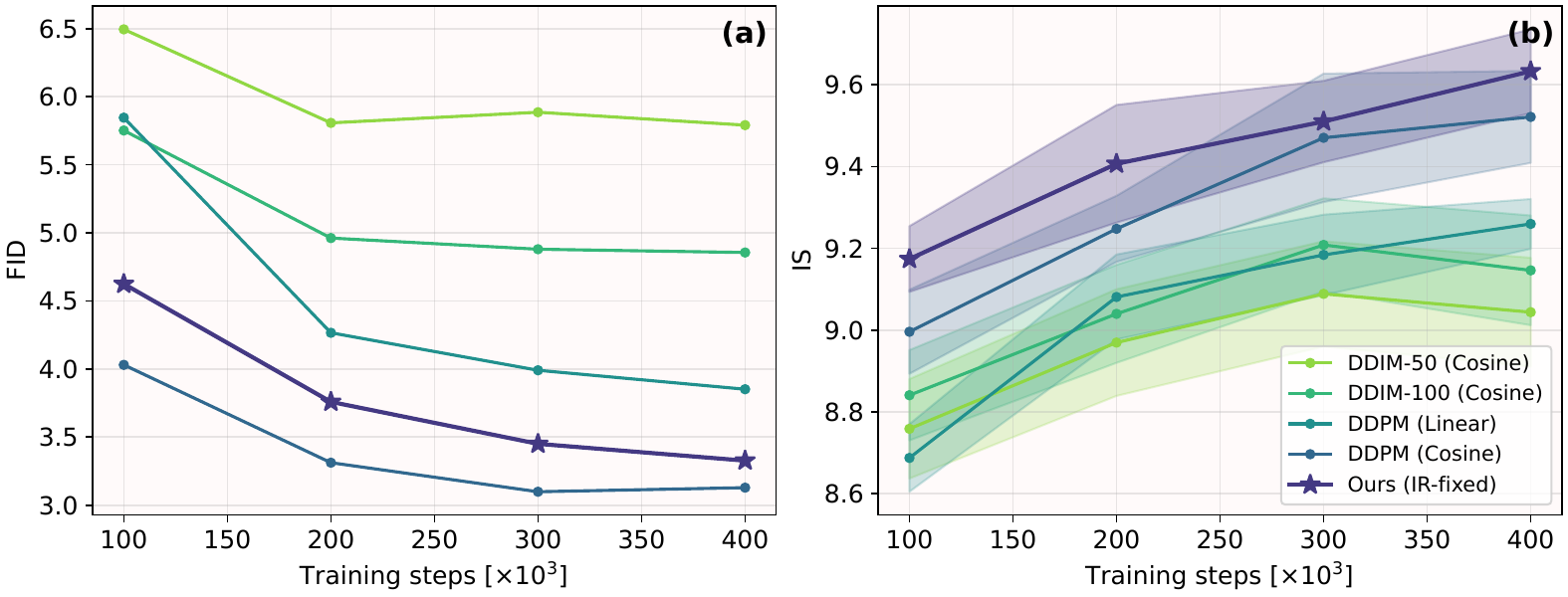}
\end{minipage}%
\hfill
\begin{minipage}{0.38\textwidth}
    \captionsetup{justification=raggedright,singlelinecheck=false}
    \caption{FID and IS convergence comparison across schedules. (a) The FID of
    our low-frequency-fixed schedule converges stably over $400$K training
    steps, outperforming the linear DDPM and two DDIM schedules. (b) The IS of
    our schedule leads the DDPM and DDIM schedules from $100$K to $400$K
    training steps.}
    \label{fig:cifar-train}
\end{minipage}
\end{figure}

Figure~\ref{fig:cifar-train} compares one of our linear schedules against
standard pixel-space DDPM and DDIM schedules. To isolate the effect of the
schedule from the mode-collapse instability discussed above, we used a
low-frequency-stable variant with $\theta = 2.0$, $\lambda_i = 137.7294$, and
$\lambda_f = 22.3$, and used the forward marginal of the ground truth as the
sampling input for diagnostics. The frequency-space schedule converges stably
over $400$K steps, with FID below the linear DDPM and the two DDIM baselines and
IS above all pixel-space schedules.

\subsubsection{Additional ablations}

\paragraph{Timestep sampler.}
The second-moment timestep sampler of \cite{dhariwal2021diffusion} reduced the
training loss faster but did not improve the final FID or IS, consistent with
low-frequency recovery being the limiting factor.

\paragraph{Number of diffusion steps.}
Reducing the number of diffusion steps from $1000$ to $500$ left FID and IS
nearly unchanged. A naive $100$-step reduction degraded both metrics. Faster
sampling is therefore feasible, but requires solver or distillation work
tailored to the mode-dependent schedule.

\paragraph{Numerical cutoffs.}
We tested numerical cutoffs of $\boldsymbol{\alpha}_n$ in $\{5\times 10^{-1},
10^{-4}, 10^{-8}\}$. The smaller cutoffs gave similar convergence with
$10^{-6}$; $5\times 10^{-1}$ degraded sample quality. This is consistent with
the low-frequency modes needing room to vary without being over-constrained.

\paragraph{Prediction target.}
Among the four prediction targets defined in Appendix~\ref{app:ddpm-theory},
$\epsilon$-prediction (used for the reported metrics) converged fastest.
Prediction of the full colored noise $\sqrt{1 - \abar_n} \odot
\sqrt{\mathbf{S}_0}\, \epsilon_n$ matched the same FID and IS but required more
training steps, while $\mathbf{X}_0$- and $\mathbf{w}$-prediction converged
slower still.

{\subsection{Generation and super-resolution from one
checkpoint}\label{app:cifar-sr}

To test both generation and super-resolution with a single model, we take a
$400$K-step CIFAR-10 generation checkpoint and run super-resolution from forward
states at effective resolutions of $2 \times 2$, $4 \times 4$, and $6 \times 6$.
This checkpoint uses the linear schedule from
Table~\ref{tab:schedule-robustness} with $k_c = 13.4$, $\lambda_i = 137.7294$,
$\lambda_f = 12$, and $\theta = 5$. Table~\ref{tab:cifar_sr} shows that FID
improves steadily as the starting state retains more low-frequency signal to
begin with.

\begin{table}[ht]
    \centering
    \small
    \caption{FID for unconditional generation and super-resolution to $32 \times
    32$ using a single CIFAR-10 checkpoint, starting from forward states at
    different effective resolutions.}
    \vspace{0.1in}
    \label{tab:cifar_sr}
    \begin{tabular}{lcccc}
    \toprule
    & Generation & $2 \to 32$ ($16\times$) & $4 \to 32$ ($8\times$) & $6 \to 32$
    ($5.3\times$) \\
    \midrule
    FID $\downarrow$ & 4.99 & 4.56 & 4.15 & 3.70 \\
    \bottomrule
    \end{tabular}
\end{table}
\par}

\section{ImageNet super-resolution experiment supplements}\label{app:imagenet}

\paragraph{Architecture and training.}
ImageNet super-resolution uses a score U-Net backbone from the NCSN++ family
\cite{song2020score} with $6$ residual blocks per resolution, attention at
resolutions $32$, $16$, and $8$, and channel multipliers $(1, 1, 2, 2, 2, 2)$.
We train with AdamW for up to $500$K steps with batch size $256$, learning rate
$10^{-4}$, weight decay $10^{-5}$, and EMA rate $0.9999$, and report metrics
from the last checkpoint. The ImageNet-$256$ model is trained on $8$ H100 GPUs;
the ImageNet-$128$ models are trained on a single H100 or GH200 GPU.

\paragraph{Effective-resolution validation.}
To compare effective-resolution forward states with conventional image resizing,
we generate a low-resolution reference by bicubic-downsampling the denormalized
image and then upsampling it to the original resolution with bicubic
interpolation (\texttt{antialias=true}). This bicubic reference validates the
SNR-defined effective-resolution interpretation used for super-resolution
experiments.

Table~\ref{tab:gt_bicubic} reports the per-dataset MSE and PSNR between the
surviving signal at each SNR threshold and the matching bicubic-degraded image.
At the threshold $\text{SNR} = 0.1$ used in all SR experiments, the MSE is
within $O(10^{-4})$ for every degradation pipeline and the PSNR exceeds $30$~dB,
indicating that the SNR-defined low-resolution input closely matches a
conventional $4\times$ or $8\times$ degradation while using the exact
forward-process marginals.

\begin{table}[t]
    \centering
    \footnotesize
    \caption{Agreement between forward-diffusion effective-resolution signals
    and bicubic down-up images across SNR thresholds. MSE is reported in units
    of $10^{-4}$ and PSNR in dB. At $\text{SNR}=0.1$, the MSE for all
    degradation pipelines is within $O(10^{-4})$ and the PSNR exceeds $30$.}
    \vspace{0.1in}
    \label{tab:gt_bicubic}
    \resizebox{\linewidth}{!}{%
    \begin{tabular}{lcccccccccccc}
    \toprule
    & \multicolumn{2}{c}{SNR=1} & \multicolumn{2}{c}{SNR=0.5} &
    \multicolumn{2}{c}{SNR=0.1} & \multicolumn{2}{c}{SNR=0.05} &
    \multicolumn{2}{c}{SNR=0.01} & \multicolumn{2}{c}{SNR=0.005} \\
    \cmidrule(lr){2-3}\cmidrule(lr){4-5}\cmidrule(lr){6-7}
    \cmidrule(lr){8-9}\cmidrule(lr){10-11}\cmidrule(lr){12-13} Dataset & MSE &
    PSNR & MSE & PSNR & MSE & PSNR & MSE & PSNR & MSE & PSNR & MSE & PSNR \\
    \midrule
    $4\times$ ImageNet-256 & 15.9 & 27.99 & 10.5 & 29.79 & 4.48 & 33.49 & 3.84 &
    34.15 & 4.53 & 33.44 & 5.33 & 32.73 \\
    $4\times$ ImageNet-128 & 21.5 & 26.67 & 14.0 & 28.53 & 4.87 & 33.12 & 3.55 &
    34.50 & 3.51 & 34.55 & 4.23 & 33.74 \\
    $8\times$ ImageNet-128 & 29.8 & 25.26 & 19.7 & 27.05 & 7.09 & 31.49 & 5.15 &
    32.88 & 4.87 & 33.12 & 5.78 & 32.38 \\
    \bottomrule
    \end{tabular}
    }
\end{table}

{
\paragraph{Starting from conventional low-resolution images.}
Our super-resolution protocol starts from the exact forward marginal of the HR
image, which carries nearly the same information as a conventional LR image. At
the $\mathrm{SNR} = 0.1$ starting point, noise dominates every frequency mode
above the effective-resolution cutoff. The noise power exceeds the signal by a
factor of $10$ at the cutoff and by roughly $10^{32}$ at the Nyquist frequency,
leaving practically no usable high-frequency detail from the HR image. Moreover, the signal
attenuation factor $e^{-\mathbf{k}^2\lambda(t)/2}$ acts as a standard Gaussian low-pass
filter.

We also verify this directly by constructing a starting state from a standard
downsampled LR image. We zero-pad the LR image in DCT space above the cutoff and
add the matching schedule noise. Table~\ref{tab:lr_state} compares this
reconstructed state against the exact forward state for $4\times$ ImageNet-256
super-resolution ($64 \to 256$), averaged across the ImageNet training set.
Across all downsamplers, the two states match within $32$ to $36$~dB in PSNR and
exceed $0.98$ in SSIM. Table~\ref{tab:gt_bicubic} provides the complementary
comparison between the noise-free forward signal and bicubic down-up images.
Pyramidal flow matching \cite{jin2025pyramidal} applies a similar noise
correction to keep different resolutions consistent within a single process.

Finally, we run end-to-end super-resolution from both starting states using a
smaller model trained for $4\times$ super-resolution ($32 \to 128$). The
resulting metrics in Table~\ref{tab:lr_sr} closely track one another, and
neither starting state dominates across all metrics.

\begin{table}[ht]
    \centering
    \small
    \caption{Comparison between the exact forward starting state and a starting
    state built from a standard LR image for $4\times$ ImageNet-256
    super-resolution ($64 \to 256$) at $\mathrm{SNR} = 0.1$. MSE is measured in
    DCT space, and all values are averaged over the ImageNet training set.}
    \vspace{0.05 in}
    \label{tab:lr_state}
    \begin{tabular}{lccc}
    \toprule
    Downsampler & MSE & PSNR (dB) & SSIM \\
    \midrule
    DCT truncation & $6.24 \times 10^{-8}$ & 36.26 & 0.9902 \\
    Lanczos & $8.10 \times 10^{-8}$ & 35.17 & 0.9883 \\
    Bicubic (anti-aliased) & $9.70 \times 10^{-8}$ & 34.38 & 0.9865 \\
    Area & $1.25 \times 10^{-7}$ & 33.24 & 0.9828 \\
    Bilinear (anti-aliased) & $1.55 \times 10^{-7}$ & 32.34 & 0.9803 \\
    \bottomrule
    \end{tabular}
\end{table}

\begin{table}[ht]
    \centering
    \small
    \caption{End-to-end $4\times$ super-resolution ($32 \to 128$) with a smaller model, starting from either the exact forward marginal or an anti-aliased bicubic LR image.}
    \vspace{0.1 in}
    \label{tab:lr_sr}
    \begin{tabular}{lcccccc}
    \toprule
    Starting state & MSE $\downarrow$ & PSNR $\uparrow$ & SSIM $\uparrow$ &
    MS-SSIM $\uparrow$ & CLIPIQA $\uparrow$ & MUSIQ $\uparrow$ \\
    \midrule
    Exact forward marginal & 0.006711 & 22.502 & 0.6356 & 0.8824 & 0.5699 &
    36.338 \\
    Bicubic LR image & 0.006697 & 22.419 & 0.6256 & 0.8790 & 0.5932 & 34.261 \\
    \bottomrule
    \end{tabular}
\end{table}
\par}

\begin{figure}[t]
    \centering
    \begin{minipage}{0.52\textwidth}
        \centering
        \includegraphics[width=\linewidth]{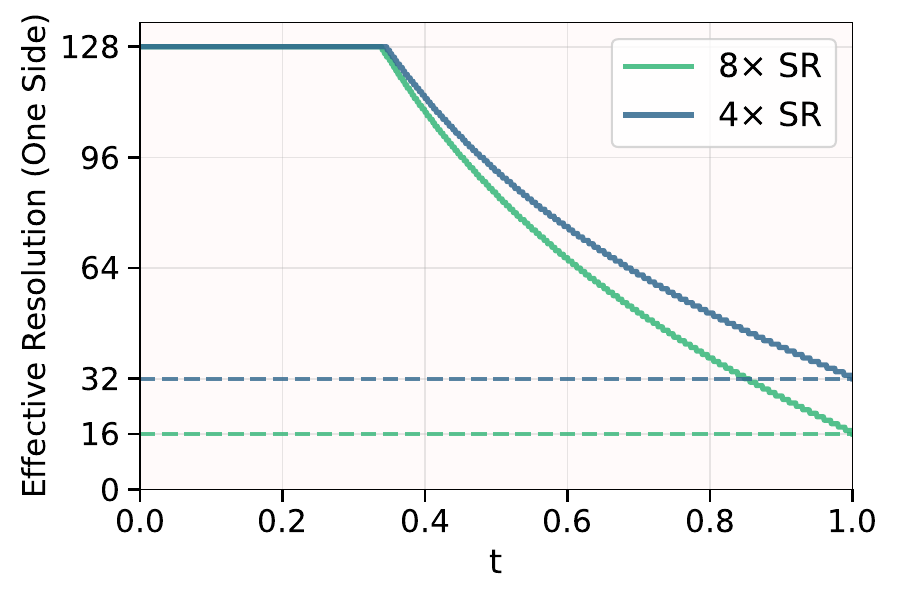}
    \end{minipage}%
    \hfill
    \begin{minipage}{0.43\textwidth}
        \captionsetup{justification=raggedright,singlelinecheck=false}
        \caption{Effective resolution against time for the ImageNet-128
        super-resolution experiments. The $8\times$ experiment ends at an
        effective resolution of $16$, and the $4\times$ experiment ends at $32$.
        For both, the maximum resolution change in a single step is $\Delta R =
        1$, so each model supports continuous super-resolution up to its
        respective final effective resolution.}
        \label{fig:res_sched}
    \end{minipage}
\end{figure}

\paragraph{Super-resolution schedules.}
The super-resolution experiments use linear schedules with $N = 1000$ steps:
\begin{align*}
    4\times\ \text{ImageNet-256:}\quad
    & \theta = 9.0,\ \lambda_i = 1132.9352,\ \lambda_f = 550.8723,\ k_c = 0; \\
    4\times\ \text{ImageNet-128:}\quad
    & \theta = 9.0,\ \lambda_i = 564.2461,\ \lambda_f = 275.4361,\ k_c = 0; \\
    8\times\ \text{ImageNet-128:}\quad
    & \theta = 5.0,\ \lambda_i = 564.2461,\ \lambda_f = 102.6489,\ k_c = 0.
\end{align*}
Figure~\ref{fig:res_sched} shows the corresponding effective-resolution paths.

{
\paragraph{Sampling cost.}
Table~\ref{tab:timing} compares the step count and per-step wall-clock time of
SKILD against diffusion super-resolution baselines, averaged over $20$ runs on a
single NVIDIA Titan RTX GPU. SinSR is distilled into a single step, while the
remaining models use multi-step diffusion samplers. Because SR3 has no official
code release, we benchmark an unofficial implementation
([https://github.com/novwaul/SR3]). SKILD has the fastest per-step runtime among
all compared methods. Its overall sampling time comes from running all $1000$
steps of the basic ancestral sampler across our experiments, as we did not
optimize sampling speed in this work. Appendix~\ref{app:sde-sample} derives the
Euler-Maruyama, probability-flow ODE, and predictor-corrector samplers for our
process, providing the foundation for fast ODE solvers and step distillation
\cite{salimans2022distillation}. Since better samplers improve both speed and
image quality in standard diffusion \cite{karras2022elucidating}, we expect
similar gains for SKILD.

\begin{table}[ht]
    \centering
    \small
    \caption{Sampling step counts and wall-clock time per step, reported as the mean and standard deviation across $20$ runs on a single NVIDIA Titan RTX GPU.}
    \vspace{0.1 in}
    \label{tab:timing}
    \begin{tabular}{lccccc}
    \toprule
    & SinSR & ResShift & I2SB & SR3 & SKILD \\
    \midrule
    Steps (NFE) & 1 & 15 & 1000 & 100 & 1000 \\
    Time per step (ms) & $1496.8 \pm 1.8$ & $222.2 \pm 1.9$ & $252.1 \pm 0.8$ &
    $144.0 \pm 0.03$ & $102.8 \pm 0.07$ \\
    \bottomrule
    \end{tabular}
\end{table}
\par}

\section{Critical Ising super-resolution details}
\label{app:ising-sr}

This appendix details the experiment in Section~\ref{sec:scientific-data}. We
generate critical Ising configurations and evaluate connected four-point
correlations across a range of scales rather than image-quality metrics.

\subsection{Data}\label{app:ising-data}

We use the ferromagnetic two-dimensional Ising model on a square lattice,
\begin{equation}
    P(s) \propto \exp\!\left(\beta \sum_{i, j \text{ being neighbors}} s_i s_j\right),
    \qquad s_i \in \{-1,+1\},
\end{equation}
at the exact critical inverse temperature
\begin{equation}
    \beta_c = \frac{1}{T_c} = \frac{1}{2}\log(1+\sqrt{2}).
\end{equation}
Configurations are sampled with the Wolff cluster algorithm
\cite{wolff1989collective}, a Markov chain Monte Carlo method for spin systems
that avoids the critical slowing down of local spin-flip dynamics. We use $128
\times 128$ binary spin fields with values in $\{-1, +1\}$ and periodic boundary
conditions.

One transition of the Wolff Markov chain constructs and flips a same-spin
cluster. For our ferromagnetic model with coupling $J = 1$, the bond probability
at criticality is
\[
    p = 1 - \exp(-2 \beta_c).
\]
A transition consists of the following steps:
\begin{enumerate}
    \item Choose a seed site uniformly at random and let $s_\star$ be its spin.
    Initialize the cluster and the active frontier to contain only this seed
    site.
    \item Given the current frontier, inspect nearest-neighbor sites with spin
    $s_\star$ that are not already in the cluster. If a candidate site touches
    $k \in \{1, 2, 3, 4\}$ frontier sites, add it to the cluster with
    probability $1 - (1 - p)^k$. This is equivalent to independently activating
    each bond from a frontier site with probability $p$.
    \item Set the newly added sites as the next frontier and repeat the previous
    step until the frontier is empty.
    \item Flip every spin in the completed cluster: $s_i \leftarrow -s_i$.
\end{enumerate}
We run several independent chains in parallel, each initialized with independent
random spins in $\{-1, +1\}^{128 \times 128}$. Every chain is thermalized for
$2{,}000$ Wolff transitions. After thermalization, we save the current
configuration from each chain whenever all of them have accumulated at least $2
L^2$ flipped spins since the last save. This is equivalent to two lattice sweeps
of cluster updates between consecutive saves.

We use $90{,}000$ samples for training and $1{,}000$ held-out samples for
testing. We do not use a separate validation set for model selection, since no
hyperparameters are tuned on the Ising benchmark. The held-out samples are
shared between SKILD and SR3, so all reported correlation differences reflect
the models and not the test inputs.

\subsection{Ground-truth forward initialization}\label{app:ising-gt-init}

We initialize the reverse process from the exact forward marginal of a
high-resolution spin field. Given $\sigma_0(\mathbf{r})$, we take its DCT to
obtain $\widehat\sigma_0(\mathbf{k})$. For a chosen timestep $n_0$, the
initialization is
\begin{equation}
    \widehat\sigma_{n_0}(\mathbf{k})
    =
    \sqrt{\abar_{n_0}(\mathbf{k})} \odot \widehat\sigma_0(\mathbf{k})
    +
    \sqrt{1 - \abar_{n_0}(\mathbf{k})} \odot
    \sqrt{\mathbf{S}_0(\mathbf{k})}\, \epsilon(\mathbf{k}),
    \qquad
    \epsilon(\mathbf{k}) \sim \mathcal{N}(0, \mathbf{I}),
    \label{eq:ising-gt-forward}
\end{equation}
which matches the forward marginal of the frequency-space DDPM used during
training. The schedule is chosen so that at $n_0 = 1000$ the low-frequency modes
corresponding to roughly a $32 \times 32$ effective resolution remain largely
intact, while higher modes are dominated by the noise term. Reverse sampling
from this initialization produces a reconstruction conditioned on the
low-frequency content of the paired held-out sample.

The diffusion uses $N = 1000$ steps with a linear schedule, $\theta = 9.0$, and
$k_c = 0$. The variance spectrum $\mathbf{S}_0$ is fit by a power law,
\begin{equation}
    \mathbf{S}_0(\mathbf{k})
    =
    C(\mathbf{k}^2 + k_0^2)^{-a},
    \qquad
    a = 0.811056,\quad
    C = 0.26641,\quad
    k_0^2 = 3.0,
\end{equation}
with $a$, $C$, and $k_0^2$ fit on the Ising training set.

\subsection{Training hyperparameters}\label{app:ising-training}

SKILD uses the same NCSN++ backbone as in the ImageNet SR experiments, with one
input and output channel, base width $128$, channel multipliers $(1, 1, 2, 2, 2,
2)$, six residual blocks per resolution, and attention at resolutions $32$,
$16$, and $8$. We train with $\epsilon$-prediction using AdamW with learning
rate $10^{-4}$, $\beta_1 = 0.9$, weight decay $10^{-5}$, $1{,}000$ warmup steps,
gradient clipping at $1.0$, batch size $256$, microbatch size $128$, and EMA
rate $0.9999$. Training uses mixed precision and a uniform timestep sampler. We
train for $100{,}000$ optimization steps and use the EMA checkpoint at this
final step for all reported results. The Ising model is trained on a single H100
or GH200 GPU.

\subsection{Connected four-point correlation}\label{app:ising-four-point}

For a fixed side length, each correlation estimate is the empirical mean over
all square patches in each image, including all translations and
symmetry-equivalent orientations, and is then averaged across images. Let
$s_{00}, s_{01}, s_{10}, s_{11}$ denote the four corner spins of a patch. The
full four-corner statistic is $G_4 = \mathbb{E}[\,s_{00}\, s_{01}\, s_{10}\,
s_{11}\,]$, the edge two-point correlation is $C_a = \mathbb{E}[\,s_{00}\,
s_{01}\,]$, and the diagonal two-point correlation is $C_b =
\mathbb{E}[\,s_{00}\, s_{11}\,]$. The connected four-point correlation reported
in the main text is
\begin{equation}
    \kappa_4 = G_4 - 2 C_a^2 - C_b^2,
    \label{eq:app-ising-kappa4}
\end{equation}
the fourth-order joint cumulant of the four corner spins. It removes the
contribution that pairwise correlations alone explain and tests whether the
model reproduces the non-Gaussian critical structure of the field. We compute
$\kappa_4$ at side lengths $\{1, 2, 4, 8, 16, 32, 64\}$.

\subsection{Paired evaluation}\label{app:ising-paired-eval}

Each generated sample is paired with the held-out spin field that produced its
low-frequency initialization. The left panel of
Figure~\ref{fig:ising-four-point} shows paired ground-truth and reconstruction
examples; the right panel reports the paired correlation comparison. The
baseline is SR3 \cite{saharia2022image}, a conditional diffusion model trained
to upsample $32 \times 32$ low-resolution Ising fields to their $128 \times 128$
ground truth.

For uncertainty estimates we use a paired bootstrap over the $1{,}000$ held-out
samples: a single bootstrap index matrix is sampled and applied to all compared
methods at every side length, preserving the sample-wise dependence induced by
conditioning on the same low-frequency input. We use $1{,}000$ bootstrap
resamples and report $99\%$ percentile confidence intervals for the curves in
Figure~\ref{fig:ising-four-point}.

\section{Additional image samples}\label{app:img_sample}

This appendix collects additional super-resolution samples on ImageNet-128 and
ImageNet-256 across the super-resolution factors reported in the main text,
along with uncurated CIFAR-10 generation samples.

\begin{figure}
    \centering
    \includegraphics[width=1.0\linewidth]{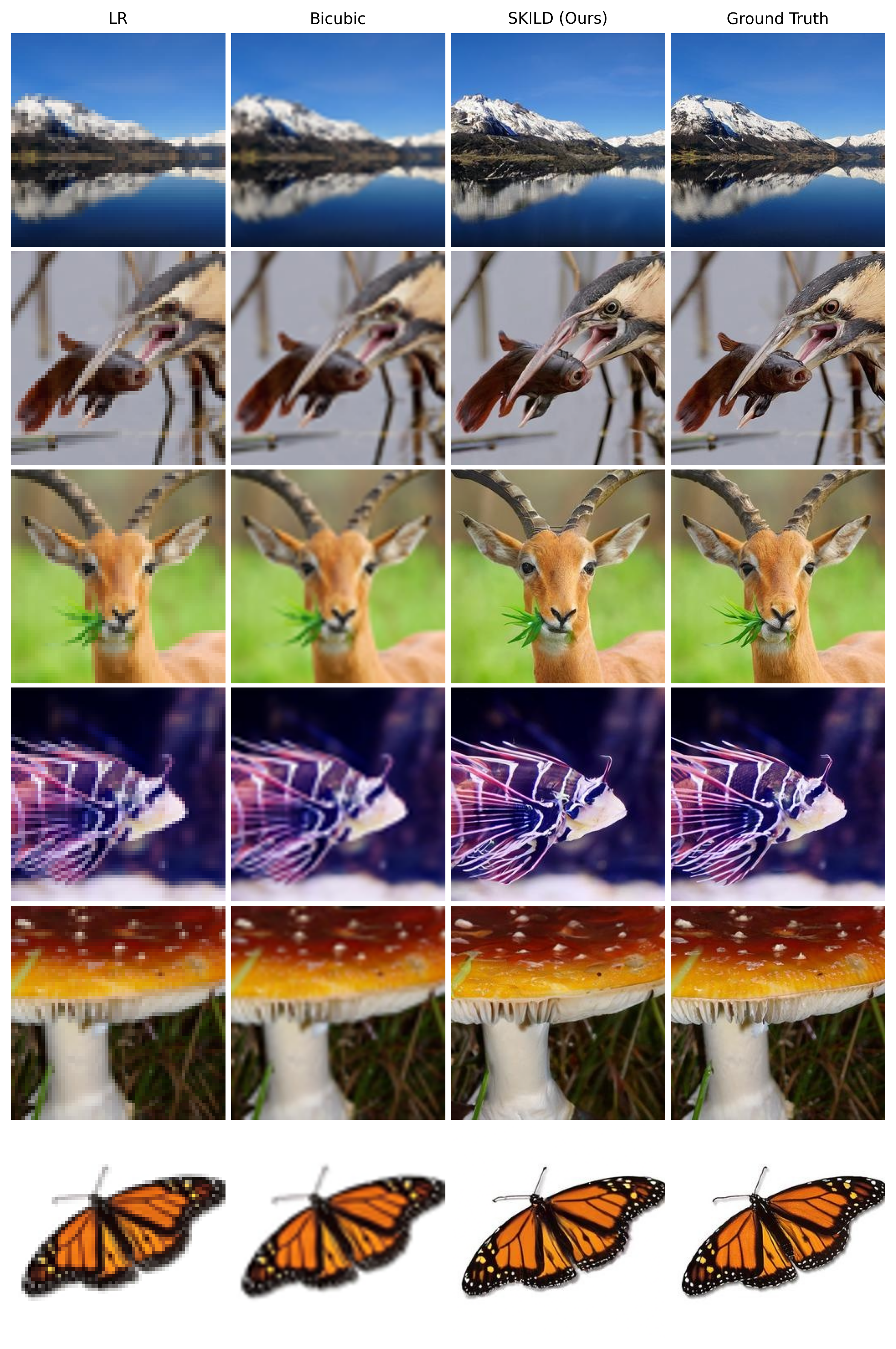}
    \caption{Additional $4\times$ super-resolution sample comparisons on
    ImageNet-256.}
    \label{fig:256_64_6}
\end{figure}

\begin{figure}
    \centering
    \includegraphics[width=1.0\linewidth]{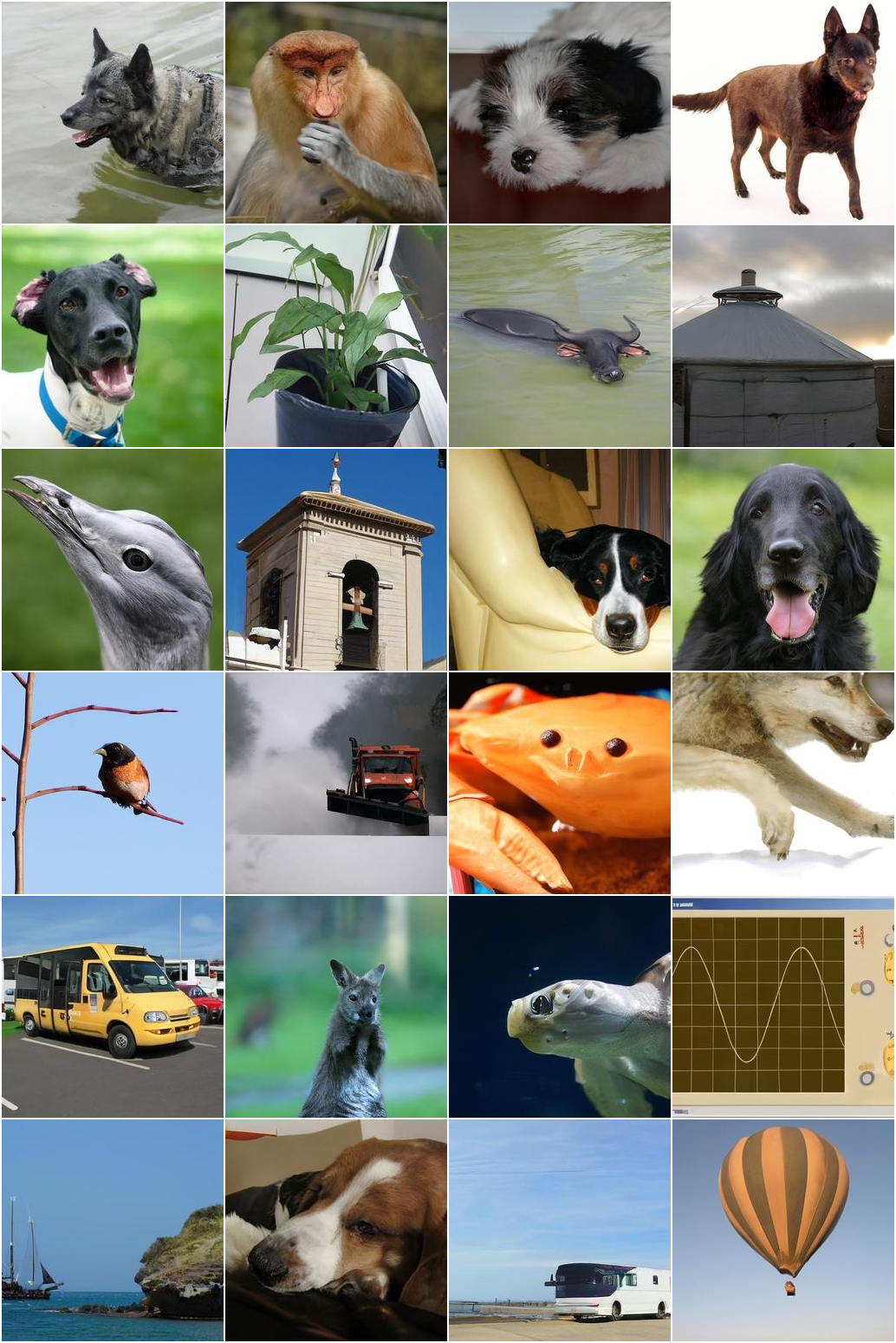}
    \caption{Additional $4\times$ super-resolution samples on ImageNet-256.}
    \label{fig:256_64_samples}
\end{figure}

\begin{figure}
    \centering
    \includegraphics[width=1.0\linewidth]{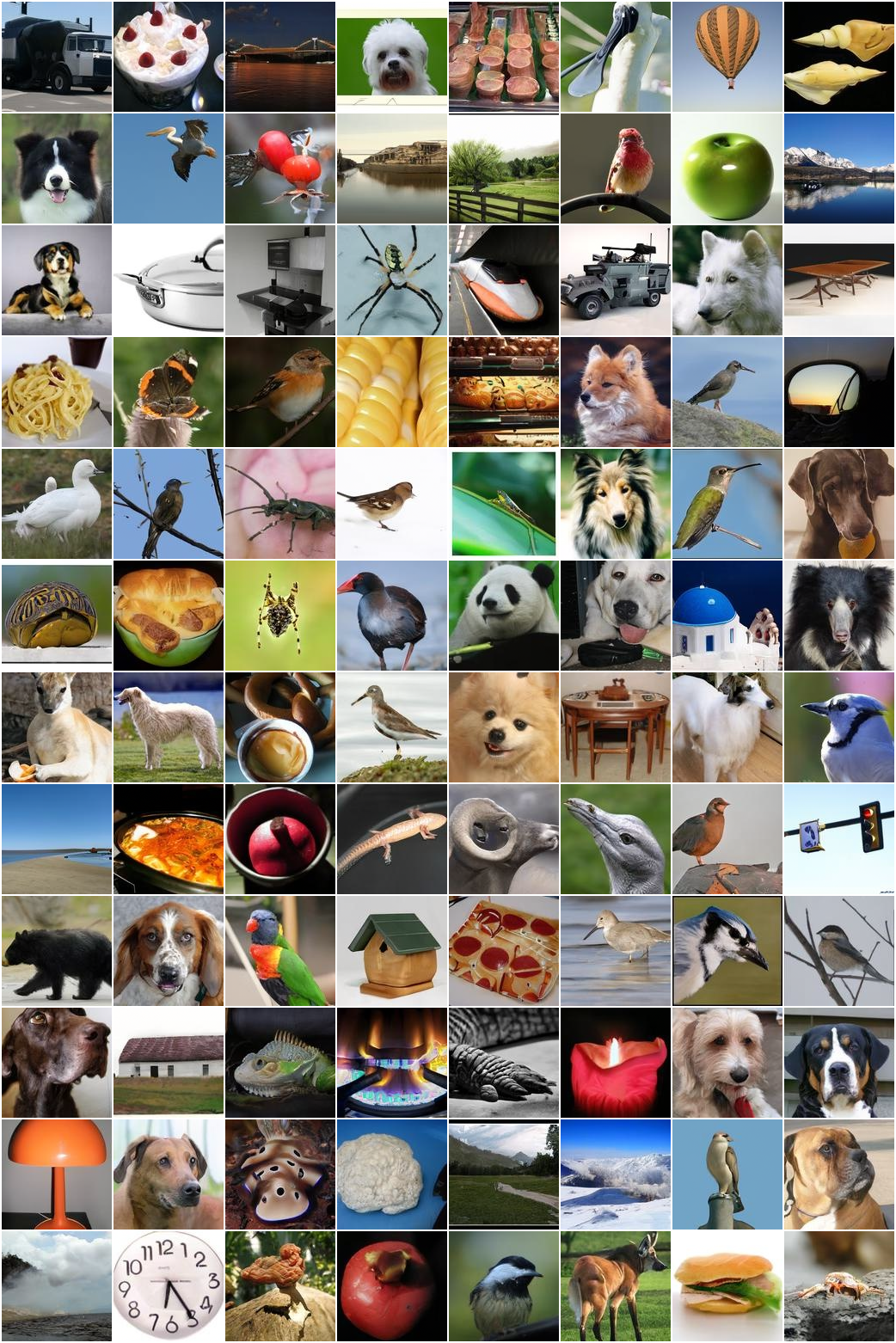}
    \caption{Additional $4\times$ super-resolution samples on ImageNet-128.}
    \label{fig:128_32_samples}
\end{figure}

\begin{figure}
    \centering
    \includegraphics[width=1.0\linewidth]{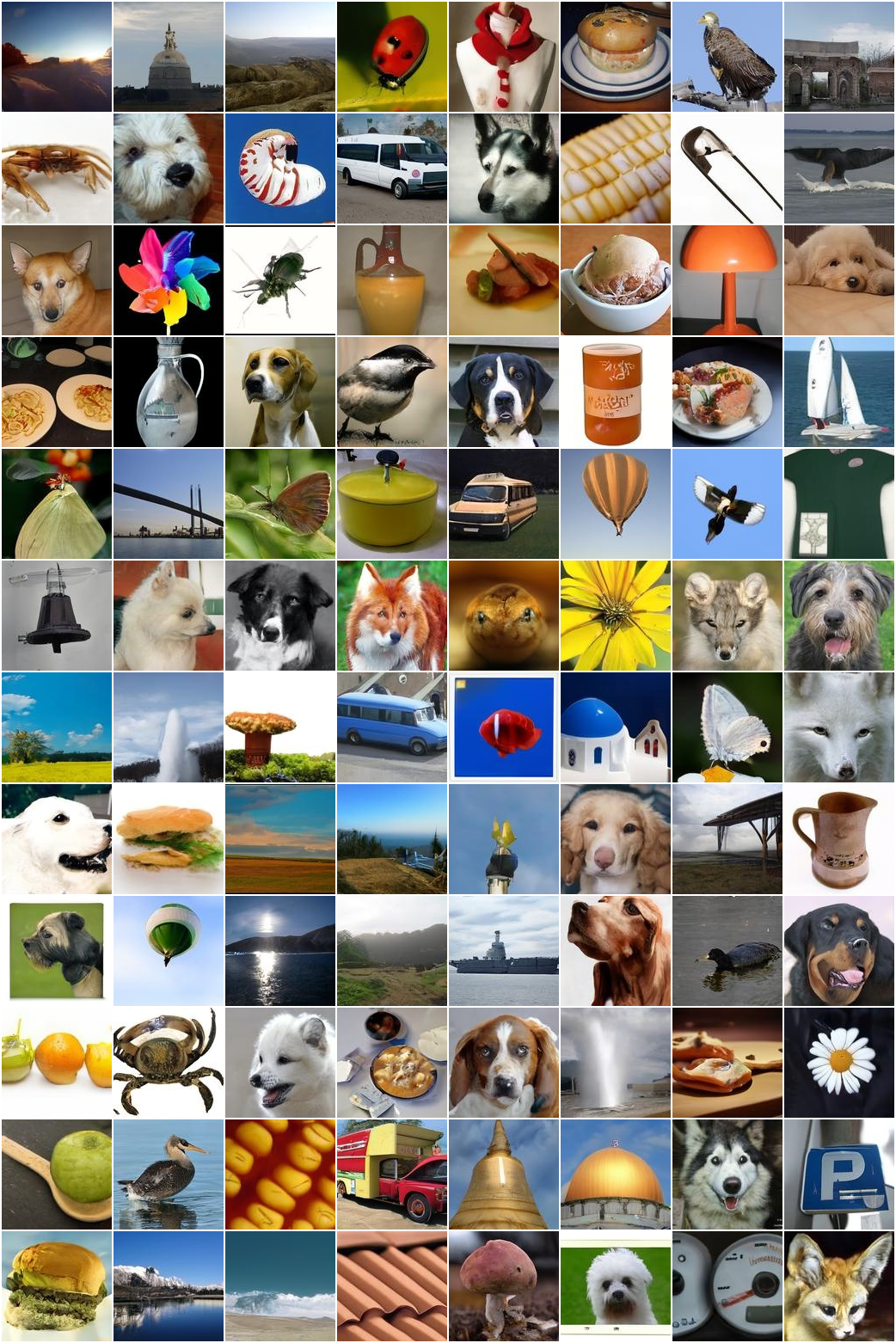}
    \caption{Additional $8\times$ super-resolution samples on ImageNet-128.}
    \label{fig:128_16_samples}
\end{figure}

\begin{figure}
    \centering
    \includegraphics[width=1.0\linewidth]{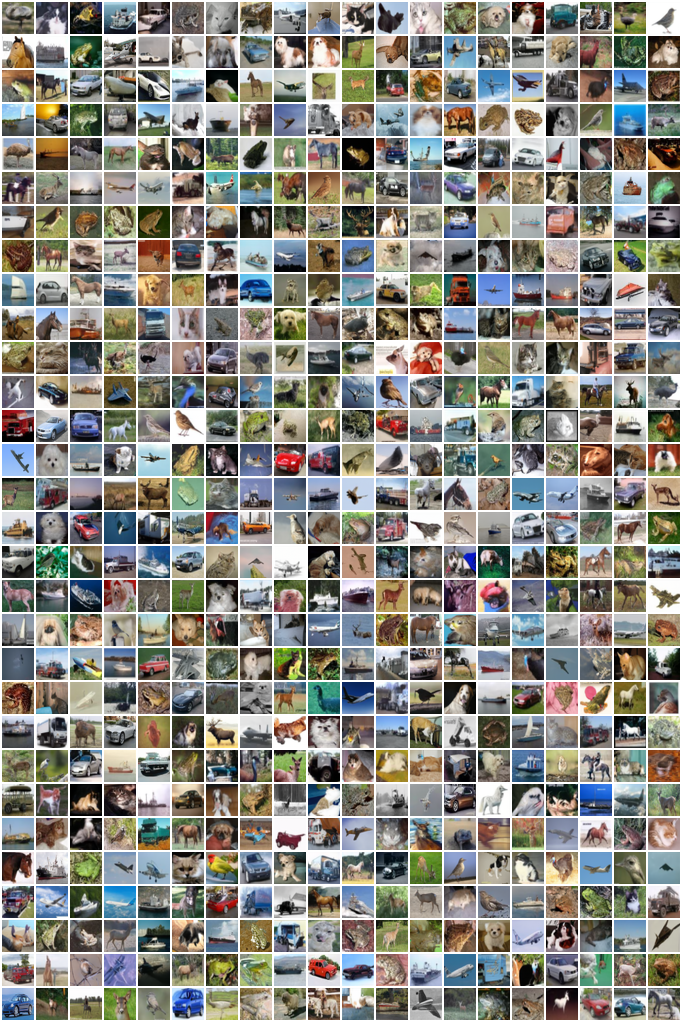}
    \caption{Uncurated samples of generated images on CIFAR-10.}
    \label{fig:cifar_600}
\end{figure}

%% file: checklist.tex
\section*{NeurIPS Paper Checklist}

\begin{enumerate}

\item {\bf Claims}
    \item[] Question: Do the main claims made in the abstract and introduction accurately reflect the paper's contributions and scope?
    \item[] Answer: \answerYes{}.
    \item[] Justification: The abstract and introduction state the method,
    assumptions, benchmark scope, and main empirical claims.

\item {\bf Limitations}
    \item[] Question: Does the paper discuss the limitations of the work performed by the authors?
    \item[] Answer: \answerYes{}.
    \item[] Justification: The Limitations section discusses sampling cost,
    low-frequency recovery, and the use of ground-truth forward marginals for super-resolution.

\item {\bf Theory assumptions and proofs}
    \item[] Question: For each theoretical result, does the paper provide the full set of assumptions and a complete (and correct) proof?
    \item[] Answer: \answerYes{}.
    \item[] Justification: The main text gives the forward marginal and DDPM
    equations, and the appendix provides the SDE and posterior derivations.

    \item {\bf Experimental result reproducibility}
    \item[] Question: Does the paper fully disclose all the information needed to reproduce the main experimental results of the paper to the extent that it affects the main claims and/or conclusions of the paper (regardless of whether the code and data are provided or not)?
    \item[] Answer: \answerYes{}.
    \item[] Justification: The paper reports datasets, model settings,
    schedules, evaluation protocols, and supplementary details needed to
    reproduce the main experiments. The code for reproducing the results are included in the supplementary material with the submission of this paper.

\item {\bf Open access to data and code}
    \item[] Question: Does the paper provide open access to the data and code, with sufficient instructions to faithfully reproduce the main experimental results, as described in supplemental material?
    \item[] Answer: \answerYes{}.
    \item[] Justification: The code for reproducing the results are included in the supplementary material with the submission of this paper. We will further post the code on GitHub following the publicaiton of the paper.

\item {\bf Experimental setting/details}
    \item[] Question: Does the paper specify all the training and test details (e.g., data splits, hyperparameters, how they were chosen, type of optimizer) necessary to understand the results?
    \item[] Answer: \answerYes{}.
    \item[] Justification: The Experiments section and appendix specify model,
    optimizer, schedule, dataset, and evaluation details.

\item {\bf Experiment statistical significance}
    \item[] Question: Does the paper report error bars suitably and correctly defined or other appropriate information about the statistical significance of the experiments?
    \item[] Answer: \answerYes{}.
    \item[] Justification: The scientific-data experiment reports bootstrap
    confidence intervals for paired correlation estimates; image-generation and
    ImageNet super-resolution metrics are reported using standard fixed-sample evaluations.

\item {\bf Experiments compute resources}
    \item[] Question: For each experiment, does the paper provide sufficient information on the computer resources (type of compute workers, memory, time of execution) needed to reproduce the experiments?
    \item[] Answer: \answerYes{}.
    \item[] Justification: The Experiments section and appendices report GPU types and
    training scales for the main experiments.
    
\item {\bf Code of ethics}
    \item[] Question: Does the research conducted in the paper conform, in every respect, with the NeurIPS Code of Ethics \url{https://neurips.cc/public/EthicsGuidelines}?
    \item[] Answer: \answerYes{}.
    \item[] Justification: The work uses standard public image datasets and
    synthetic scientific data generation, and does not involve human subjects or
    sensitive personal data.

\item {\bf Broader impacts}
    \item[] Question: Does the paper discuss both potential positive societal impacts and negative societal impacts of the work performed?
    \item[] Answer: \answerYes{}.
    \item[] Justification: The Limitations and broader-impact discussion cover
    benefits for multi-scale modeling and risks of plausible but incorrect
    super-resolution details.
    
\item {\bf Safeguards}
    \item[] Question: Does the paper describe safeguards that have been put in place for responsible release of data or models that have a high risk for misuse (e.g., pre-trained language models, image generators, or scraped datasets)?
    \item[] Answer: \answerNA{}.
    \item[] Justification: The paper does not release a high-risk foundation
    model or scraped dataset; any released code and scientific data will be
    accompanied by usage notes.

\item {\bf Licenses for existing assets}
    \item[] Question: Are the creators or original owners of assets (e.g., code, data, models), used in the paper, properly credited and are the license and terms of use explicitly mentioned and properly respected?
    \item[] Answer: \answerYes{}.
    \item[] Justification: The paper cites the external datasets, baselines, and
    methods used in the experiments.

\item {\bf New assets}
    \item[] Question: Are new assets introduced in the paper well documented and is the documentation provided alongside the assets?
    \item[] Answer: \answerYes{}.
    \item[] Justification: The scientific benchmark is documented in the main
    text and appendix, including generation and evaluation details.

\item {\bf Crowdsourcing and research with human subjects}
    \item[] Question: For crowdsourcing experiments and research with human subjects, does the paper include the full text of instructions given to participants and screenshots, if applicable, as well as details about compensation (if any)? 
    \item[] Answer: \answerNA{}.
    \item[] Justification: The work does not involve crowdsourcing or research
    with human subjects.

\item {\bf Institutional review board (IRB) approvals or equivalent for research with human subjects}
    \item[] Question: Does the paper describe potential risks incurred by study participants, whether such risks were disclosed to the subjects, and whether Institutional Review Board (IRB) approvals (or an equivalent approval/review based on the requirements of your country or institution) were obtained?
    \item[] Answer: \answerNA{}.
    \item[] Justification: The work does not involve human subjects, so IRB or
    equivalent approval is not applicable.

\item {\bf Declaration of LLM usage}
    \item[] Question: Does the paper describe the usage of LLMs if it is an important, original, or non-standard component of the core methods in this research? Note that if the LLM is used only for writing, editing, or formatting purposes and does \emph{not} impact the core methodology, scientific rigor, or originality of the research, declaration is not required.
    \item[] Answer: \answerNA{}.
    \item[] Justification: LLMs are not part of the core methodology,
    experiments, or scientific claims.

\end{enumerate}